\documentclass[pdflatex,sn-mathphys-num]{sn-jnl}% Math and Physical Sciences Numbered Reference Style
\usepackage{graphicx}%
\usepackage{multirow}%
\usepackage{amsmath,amssymb,amsfonts}%
\usepackage{amsthm}%
\usepackage{mathrsfs}%
\usepackage[title]{appendix}%
\usepackage{xcolor}%
\usepackage{textcomp}%
\usepackage{manyfoot}%
\usepackage{booktabs}%
\usepackage{array}
\usepackage{algorithm}%
\usepackage{algorithmicx}%
\usepackage{algpseudocode}%
\usepackage{listings}
\usepackage[utf8]{inputenc}
\usepackage{array}%
\usepackage{multirow}
\usepackage{bm}
\usepackage{subcaption} 
\usepackage{float}   
\usepackage[export]{adjustbox}
\usepackage{comment}
\theoremstyle{thmstyleone}%
\theoremstyle{thmstyletwo}%

\theoremstyle{thmstylethree}%

\begin{document}

\title[Genome-to-Image Translation of \textit{E. coli} Morphology]{Genome-to-Image Translation:
Reconstructing \textit{Escherichia coli} Morphology from Genomic Sequences}


\author[1,2,3]{\fnm{Son T.} \sur{Huynh}}\email{htson@hcmus.edu.vn}
\equalcont{These authors contributed equally to this work.}

\author[1,2,3]{\fnm{Thuy T.} \sur{Le}}
\equalcont{These authors contributed equally to this work.}

\author[1,2,3]{\fnm{Thu H. A.} \sur{Huynh}}
\equalcont{These authors contributed equally to this work.}

\author*[1,2,3]{\fnm{Binh T.} \sur{Nguyen}}\email{ntbinh@hcmus.edu.vn}

\affil*[1]{\orgname{AISIA Research Lab}, \orgaddress{\city{Ho Chi Minh City}, \country{Vietnam}}}

\affil[2]{\orgdiv{Faculty of Mathematics and Computer Science}, \orgname{University of Science, Ho Chi Minh City}, \orgaddress{\country{Vietnam}}}

\affil[3]{\orgname{Vietnam National University, Ho Chi Minh City}, \orgaddress{\country{Vietnam}}}


\abstract{Antimicrobial resistance in \textit{Escherichia coli} is established in
routine practice by culture-based susceptibility testing, which is slow relative
to the clinical need for an early treatment decision. Resistance status also
leaves a measurable trace in cell morphology, and whole-genome sequencing of
clinical isolates is increasingly available, which raises the question of whether
colony appearance can be predicted directly from an isolate's genome. We
formulate this as a conditional image-generation problem. The assembled contigs
of each isolate are encoded into a single 768-dimensional vector by a frozen
genomic language model through a segment--contig--isolate aggregation hierarchy,
and a conditional Wasserstein GAN with gradient penalty synthesizes colony images
from this embedding together with the applied antibiotic and the resistance
label. We train on 167,356 single-cell images from 22 clinical blood isolates
imaged under azithromycin, ciprofloxacin, tetracycline, and a no-drug control,
compare seven pretrained convolutional backbones inside the generator, and select
ConvNeXt-Tiny on distributional image-quality metrics. The synthetic images are
then assessed extrinsically: added to a reduced real training set, they improve
per-antibiotic resistance classification, with a significant gain in
resistant-class recall for all three antibiotics when isolates are shared across
splits, and smaller drug-dependent gains when test isolates are held out
entirely. Gains concentrate in whichever class the baseline classifier handles
worst.}

\keywords{Antimicrobial resistance, Genome-to-image translation,
Conditional GAN, Genomic language model, Data augmentation,
\textit{Escherichia coli}}



\maketitle

\section{Introduction}

Antimicrobial resistance limits the treatment options available for common
bacterial infections, and \textit{Escherichia coli} bloodstream infection is
among the settings where the cost of an ineffective first-line choice is
highest~\cite{murray2022global,who2022glass}. In routine practice the resistance profile of an isolate is
established by culture-based antimicrobial susceptibility testing, which
requires the organism to grow under each candidate drug before a phenotype can
be read out. The resulting delay is incompatible with the clinical need to
commit to a therapy early, so the initial prescription is frequently empirical
and is revised once susceptibility results arrive~\cite{vanbelkum2020rapid}.

Two developments make it reasonable to ask whether this phenotype can be
anticipated computationally. The first is that drug exposure and resistance
status are visible in the cell itself: susceptible and resistant
\textit{E.~coli} differ in subcellular morphology under antibiotic challenge
strongly enough that a convolutional classifier can separate them from
microscopy alone~\cite{Zagajewski2023-to}. The second is that whole-genome
sequencing of clinical isolates has become routine, and genomic language models
pretrained on large nucleotide corpora now provide general-purpose
representations of raw sequence without task-specific
feature engineering~\cite{zhou2024dnabert}. Taken together, these suggest that
the genome of an isolate carries information about how that isolate will
\emph{look} under a given drug, not only about which resistance determinants it
carries.

Existing work does not connect these two ends. Genotype-to-phenotype prediction
has overwhelmingly treated the phenotype as a scalar or structured numerical
target, including genome-based prediction of resistance
itself~\cite{su2019genome}, so the visual morphology of the organism is never
modelled. Conditional
image generation from molecular descriptors, meanwhile, has been developed for
eukaryotic systems and is conditioned on processed readouts such as gene
expression, chemical structure, or a discrete perturbation
label~\cite{lee2022morphnet,yang2021mol2image,palma2025impa,navidi2025morphodiff,cai2025ge2hist}.
The one method we are aware of that generates morphological images directly from
DNA, G2PDiffusion~\cite{liu2025g2pdiffusion}, targets morphological diversity
across insect taxa using multiple sequence alignments and geographic metadata,
and has no notion of drug treatment or resistance. To our knowledge, no prior
method conditions image generation on an embedding learned from raw bacterial
sequencing reads, nor relates the generated images to an antimicrobial
resistance phenotype.

This paper addresses that gap. We cast the problem as genome-to-image
translation for a single clinically relevant species and build a conditional
generative framework with two parts. A genomic encoding pipeline takes the raw
paired-end reads of an isolate through quality control, merging, assembly, and
annotation, then encodes the resulting contigs into one fixed-length
768-dimensional vector with a frozen DNABERT-2 encoder, aggregating
hierarchically from segments to contigs to the isolate with a length-aware
weighted average. A conditional Wasserstein GAN with gradient penalty
(cWGAN-GP)~\cite{gulrajani2017improved} then synthesizes colony images from this
embedding together with the applied antibiotic and, in the second training
phase, the resistance status of the isolate. The adversarial formulation with
gradient penalty is chosen deliberately over a diffusion backbone: our dataset
comprises a single species represented by 22 isolates, a regime in which a
data-intensive generative model is difficult to justify.

The setting also shapes how we evaluate. Intrinsic generative metrics report
whether synthetic images resemble real ones distributionally, but they do not
establish that the images are biologically informative. We therefore pair FID,
KID, and SSIM with a task-based benchmark: synthetic cells conditioned on
antibiotic and resistance status are added to a deliberately reduced real
training set, and we measure what they do to a downstream per-antibiotic
resistance classifier across five seeds with paired significance testing, under
both a permissive split and a stricter split in which no test isolate is seen
during training.

Our contributions are as follows.
\begin{itemize}
  \item We formulate bacterial genome-to-image translation conditioned on a
        genomic foundation-model embedding, the applied antibiotic, and the
        resistance phenotype, which to our knowledge has not been attempted for
        bacterial morphology or antimicrobial resistance.
  \item We describe a reproducible genomic encoding pipeline that maps raw
        paired-end reads to a single isolate-level embedding, including an
        adaptive read-merging threshold derived per isolate from the observed
        insert-size distribution rather than a fixed cutoff.
  \item We report a controlled ablation of seven ImageNet-pretrained CNN
        backbones inside the generator under identical training conditions, and
        document two concrete training failure modes of the naive formulation
        together with the objective design that resolves them.
  \item We evaluate the synthetic images extrinsically rather than only
        intrinsically, showing that GAN augmentation improves resistant-class
        recall under a low-data regime and that the gains concentrate in the
        class the baseline classifier handles worst, while being explicit about
        the accompanying precision trade-off and about the limits of training
        the generator and the classifier on the same dataset.
\end{itemize}

The remainder of the paper is organized as follows.
Section~\ref{sec:related} reviews molecularly conditioned image generation and
morphology-based resistance characterization. Section~\ref{sec:dataset}
describes the paired sequencing and imaging data and the splitting protocols.
Section~\ref{sec:method} presents the genomic encoding pipeline and the
cWGAN-GP framework. Section~\ref{sec:experiments} reports the backbone ablation
and the downstream classification results, and
Section~\ref{sec:conclusion} concludes.

\section{Related Work}\label{sec:related}

The most closely related line of work generates cellular images from
molecular descriptors, motivated largely by phenotypic drug discovery and
high-content microscopy. We review it in detail because it shares our core
setting, in which a generator is conditioned on a molecular representation
to synthesize a morphological image.

An early instance is MorphNet~\cite{lee2022morphnet}, which learns the
conditional distribution of cell morphology given single-cell gene
expression by pairing a variational autoencoder that compresses the
expression vector with a generative adversarial network that decodes the
resulting representation into an image. Mol2Image~\cite{yang2021mol2image}
instead conditions a flow-based generative model on the chemical structure
of a compound, synthesizing cell microscopy images for unseen molecular
interventions through a multi-scale architecture trained with a contrastive
objective. More recent methods scale this idea and improve generalization.
IMPA~\cite{palma2025impa} casts perturbation prediction as a style-transfer
problem, using an autoencoder-based conditional adversarial model to predict
how a control cell would appear under a given chemical or genetic
intervention, and demonstrates extrapolation to perturbations not seen
during training on breast cancer and osteosarcoma cell lines.
MorphoDiff~\cite{navidi2025morphodiff} integrates perturbation embeddings as
guiding signals within a latent diffusion model and produces high-resolution
morphological predictions that generalize across both chemical and genetic
interventions on Cell Painting datasets. A parallel thread targets histopathology, where the molecular signal is
gene expression and the image is a tissue section.
GE2Hist~\cite{cai2025ge2hist} derives cell-state representations from
single-cell gene expression using a single-cell foundation model and
conditions a diffusion model to synthesize histological images that
reconstruct cellular composition and spatial organization. Related efforts
extend this multimodal coupling to jointly predict morphology and
expression and to adapt pretrained diffusion models to data-limited
institutional settings, reflecting growing interest in molecularly
conditioned image synthesis for tissue.

Two observations follow from this body of work, and together they motivate
the present study. First, the molecular conditioning signal in all of these
methods is a processed, eukaryotic readout, namely gene expression,
chemical structure, or a discrete perturbation label, rather than a
representation derived directly from raw genomic sequence. Second, the
biological targets are exclusively eukaryotic, comprising cultured mammalian
cell lines and human tissue sections, with no treatment of bacterial
morphology or the clinical question of antimicrobial resistance. To our
knowledge, no prior method conditions image generation on an embedding
learned directly from raw bacterial sequencing reads, nor links the
generated images to antimicrobial resistance phenotypes, which is precisely
the gap this work addresses.

%G2PDiffusion
Predicting phenotype from genotype is a long-standing problem, but most
computational approaches predict scalar or structured numerical traits
rather than images, treating the phenotype as a low-dimensional regression
or classification target and therefore never modelling the visual
morphology of the organism. Recasting genotype-to-phenotype prediction as
\emph{image generation} is recent, and to the best of our knowledge
G2PDiffusion~\cite{liu2025g2pdiffusion} is the only existing method that
generates morphological images directly from genomic sequence. It treats
images as morphological proxies and conditions a diffusion model on a
genotype-environment representation built from two evolutionary signals: a
multiple sequence alignment (MSA), retrieved from an external database with
MMseqs2 to capture conserved and co-evolutionary regions, and geographic
coordinates encoding environmental context. Trained on the BIOSCAN-5M
dataset of over five million insect specimens paired with DNA barcodes,
coordinates, and images, it is evaluated against adapted conditional
generation baselines using alignment- and embedding-based measures of
genotype-phenotype consistency.

Although G2PDiffusion and our work share the high-level idea of generating
images from DNA, they differ in ways that matter for the antimicrobial
resistance setting. G2PDiffusion studies morphological diversity across
diverse insect taxa and has no notion of drug treatment or resistance,
whereas our generator targets a single clinically relevant species,
\textit{E. coli}, and is conditioned on the applied antibiotic and a
binary resistance label so that the synthesized colony morphology reflects a
specific drug exposure and resistance state. Their conditioning signal is an
MSA assembled by cross-species homology retrieval augmented with geographic
metadata, whereas ours is a fixed-length isolate-level embedding produced
directly by a genomic foundation model from each isolate's assembled reads,
requiring neither homology retrieval nor environmental coordinates. The two
settings also occupy opposite data regimes: G2PDiffusion is a diffusion
model trained on more than five million DNA-image pairs, while our setting
comprises a single species represented by only 22 isolates, a scarcity that
partly motivates our choice of an adversarial objective with gradient
penalty over a data-intensive diffusion backbone. Finally, G2PDiffusion is evaluated
intrinsically, through DNA-image alignment and phenotype embedding
similarity, whereas our central evaluation is extrinsic: synthetic images
augment a downstream resistance classifier whose per-antibiotic performance
is assessed with explicit significance testing. To our knowledge, no prior genome-to-image
method addresses bacterial morphology or antimicrobial resistance, which is
the gap this work targets.

A separate line of work establishes that antimicrobial susceptibility
leaves a visible morphological signature in bacterial cells, and exploits
this signal discriminatively. Most directly related is
DASP~\cite{Zagajewski2023-to}, which images fluorescently stained
\textit{E.~coli} cells and trains convolutional networks to classify
individual cells as susceptible or resistant from subcellular morphology,
across four antibiotics with distinct mechanisms of action including
ciprofloxacin, and validates the approach on clinical isolates. This
confirms the premise underlying our work, namely that resistance status is
encoded in cellular appearance, but it is a purely discriminative
method: it classifies existing images rather than generating them, and it
is conditioned on the image alone, with no genomic input. Generative models
have separately been applied to bacterial imaging, but not for resistance
characterization. Specifically, CycleGAN-based translation has been used to
turn procedurally generated colony images into realistic microscopy for
training segmentation models~\cite{10.1371/journal.pcbi.1012874}, and related
efforts synthesize bacterial or biofilm imagery to enlarge training sets or
emulate staining. In all of these cases the generator is conditioned on
image-derived or procedural signals rather than on genomic information, and
none addresses antimicrobial resistance. Our work differs from both threads: rather than classifying images
discriminatively or generating images from image-derived cues, we generate
resistance-conditioned colony images from a genomic embedding and use them to
improve a downstream resistance classifier. In doing so, this work contributes
an application of conditional generative modeling to a biological imaging
domain, extending GenAI-based synthesis beyond the natural-image and
human-communication settings in which it is most often studied toward the
simulation of microbiological systems from molecular descriptors. It also
contributes on the evaluation axis: beyond reporting standard intrinsic
generative metrics (FID, KID, SSIM), we adopt a task-based benchmark in which
the synthesized images are assessed by their measurable utility for a
downstream resistance-classification task, with per-antibiotic performance
quantified across multiple seeds under explicit significance testing. This
pairing of intrinsic image-quality metrics with an extrinsic, task-grounded
benchmark speaks directly to the need for rigorous evaluation protocols for
generative models in applied, high-stakes domains.

\section{Dataset}\label{sec:dataset}
\subsection{Data Acquisition}

This study pairs two modalities acquired from the same collection of bacterial
isolates: whole-genome sequencing of each isolate, and colony microscopy of that
isolate imaged under an antibiotic panel. Both modalities are keyed by isolate
identifier, which is what allows a genomic representation to be used as the
conditioning signal for an image of the same organism.

\paragraph{Isolates and sequencing.}
The genomic material originates from the
19EN\_single\_isolate\_fromBlood collection, comprising 27 single-isolate
samples recovered from blood specimens and used here for species identification
and for genomic encoding. Each isolate was sequenced under a paired-end
protocol, and the reads of an isolate were archived as two gzip-compressed
FASTQ files corresponding to the forward and reverse mates
(\texttt{R1.fastq.gz} and \texttt{R2.fastq.gz}). Read quality, adapter content,
and insert-size distribution were assessed per isolate as part of the
preprocessing pipeline described in Section~\ref{sec:genomic}; no isolate was
excluded on the basis of sequencing quality.

\paragraph{Imaging.}
Colony microscopy was performed in three sessions on three separate dates,
yielding the three imaging disks summarized in Table~\ref{tab:disk_data}, with
nine isolates imaged per disk. On every disk each isolate was imaged under the
full panel of four conditions: azithromycin (AZI), ciprofloxacin (CIP),
tetracycline (TE), and an untreated no-drug control, so that the morphological
effect of a drug can be read against the untreated appearance of the same
isolate. The ATCC~25922 reference strain of \textit{E.~coli} was imaged
alongside the clinical isolates as a quality control for the imaging protocol;
because no sequencing data exist for it, it carries no genomic embedding and is
excluded from all modelling. Acquisition produced 549 microscopy images in
total, from which 173,081 individual cell images were obtained by segmenting and
cropping each detected cell to its own image. All experiments in this paper
operate on these single-cell crops rather than on whole microscopy fields.

\paragraph{Resistance labelling.}
Every isolate--antibiotic pair carries a binary ground-truth phenotype,
resistant (R) or susceptible (S), determined by antimicrobial susceptibility
testing independently of the imaging. These labels are the targets of the
downstream classification task in Section~\ref{sec:performance} and the
resistance conditioning signal of the Phase~2 generator. Labels are defined only
for the three antibiotic-treated conditions; no-drug control images carry no
resistance label and are therefore excluded wherever the resistance phenotype is
required.

\subsection{Data Overview}
The dataset used in this study consists of two complementary data modalities: whole-genome sequencing data and microscopy image data, both derived from bacterial isolates.

\subsubsection{Genome Sequencing data} 
We use genome sequencing from the 19EN\_single\_isolate\_fromBlood dataset, comprising 27 isolates collected from blood samples, for species identification. For each isolate, paired-end reads were stored as R1/R2 FASTQ files (R1/R2.fastq.gz).

\subsubsection{Image data}
Colony microscopy images are collected across three imaging sessions conducted on three separate dates: disk 1 (22/12/2021), disk 2 (30/12/2021), and disk 3 (01/01/2022); each disk covers a distinct set of nine isolates. On each disk, every isolate is imaged under a full antibiotic panel comprising four treatment conditions: azithromycin (AZI), ciprofloxacin (CIP), tetracycline (TE), and a no-drug control. The per-disk statistics are summarized in Table~\ref{tab:disk_data}.

\begin{table}[htbp]
\centering
\caption{Statistics on the number of Isolates, Images, and Cells per Disk}
\label{tab:disk_data}
\begin{tabular}{ccccc}
\toprule
\textbf{Disk} & \textbf{Date} & \textbf{Isolates} & \textbf{Images} & \textbf{Cells} \\ \midrule
1 & 22/12/2021 & 9 & 213 & 74,359 \\
2 & 30/12/2021 & 9 & 211 & 37,502 \\
3 & 01/01/2022 & 9 & 125 & 61,220 \\ \midrule
\textbf{Total} & & & \textbf{549} & \textbf{173,081} \\ \bottomrule
\end{tabular}
\end{table}

Among the 549 images, 29 images (5,725 cells) correspond to the ATCC 25922 reference strain, which lacks genomic sequence data and is therefore excluded from downstream analysis. The remaining 520 images (167,356 cells) represent 22 unique clinical isolates, all of blood origin, which constitute the working dataset for this study.

\subsubsection{Data splitting}
To accommodate the distinct objectives and metadata criteria of each research phase, the data partitioning strategy for training, validation, and testing sets varies across phases, as detailed below.

\paragraph{Phase 1 (cWGAN-GP)}
For the generative modeling phase, the dataset was partitioned at the single-cell level into training, validation, and testing sets with a 70:20:10 ratio. Cell-level splitting is adopted because several isolate–antibiotic combinations contain only a single image, making image-level partitioning infeasible for meaningful evaluation. To preserve experimental diversity, each split is stratified to ensure all 22 isolates and all four treatment conditions are represented, and the 70:20:10 ratio is enforced within every isolate–antibiotic subgroup. The resulting partition yields 117,106 cells for training, 33,473 for validation, and 16,777 for testing (total: 167,356 cells).

\paragraph{Phase 2 (AMR classification)}
For the downstream resistance/susceptibility classification task, two data-partitioning strategies were considered, both performed at the image level into training, validation, and testing sets to prevent data leakage. Since a single microscopy image can contain hundreds of individual bacterial cell images, these cell images may share the same resistance/susceptibility label, which would undermine the reliability of evaluation on the test set if they were scattered randomly across different splits. Accordingly, all cell images originating from the same microscopy image are assigned exclusively to a single data split. Only images from the three antibiotic-treated conditions (AZI,~CIP,~TE) are included; no-drug control images are excluded as they carry no resistance or susceptibility label and are therefore not applicable to the binary classification objective. The per-antibiotic split statistics for the two strategies are reported in Tables~\ref{tab:data_agnostic} and~\ref{tab:data_aware}.

\textbf{Isolate-agnostic splitting.} In the first strategy, isolate identity is not taken into account: microscopy images from all isolates are pooled and randomly partitioned into training, validation, and testing sets at the image level. As a result, images from the same isolate may appear in different splits, although cell images from the same microscopy image are still never separated. This setting is less stringent and measures performance when the model has already encountered the isolates during training.

\begin{table}[htbp]
\centering
\caption{Number of images and cells per split under the \textbf{isolate-agnostic} splitting strategy.}
\label{tab:data_agnostic}

\begin{tabular}{clcccccc}
\toprule
\multirow{2}{*}{\textbf{Antibiotic}} & \multirow{2}{*}{\textbf{Split}} 
& \multicolumn{2}{c}{\textbf{Total}} & \multicolumn{2}{c}{\textbf{R Label}} & \multicolumn{2}{c}{\textbf{\% R}} \\ 
& & Images & Cells & Images & Cells & Images & Cells \\
\midrule

\multirow{3}{*}{AZI} & Train & 78 & 26,967 & 57 & 19,002 & 73\% & 70\% \\
                     & Val   & 22 & 4,935  & 14 & 3,191  & 64\% & 65\% \\
                     & Test  & 38 & 6,561  & 29 & 4,873  & 76\% & 74\% \\ \midrule
\multirow{3}{*}{TE}  & Train & 68 & 31,220 & 45 & 21,022 & 66\% & 67\% \\
                     & Val   & 22 & 6,795  & 16 & 4,779  & 73\% & 70\% \\
                     & Test  & 33 & 6,700  & 20 & 3,864  & 61\% & 58\% \\ \midrule
\multirow{3}{*}{CIP} & Train & 79 & 24,378 & 47 & 17,021 & 59\% & 70\% \\
                     & Val   & 22 & 4,716  & 14 & 3,340  & 64\% & 71\% \\
                     & Test  & 41 & 5,814  & 22 & 3,765  & 54\% & 65\% \\ \bottomrule
\end{tabular}
\end{table}

\textbf{Isolate-aware splitting.} In the second strategy, the training and testing sets are separated at the isolate level, so that no isolate appears in both sets. Specifically, $\sim$80\% of the isolates are assigned to the training set and the remaining isolates form the testing set. The validation set is then drawn from the training set, accounting for $\sim$25\% of its microscopy images, and is split at the image level. Consequently, an isolate belonging to the training portion may contribute images to both the training and validation sets, but never to the testing set. This strategy evaluates the model's ability to generalize to previously unseen isolates, which more closely reflects real-world deployment.

\begin{table}[htbp]
\centering
\caption{Number of images and cells per split under the \textbf{isolate-aware} splitting strategy.}
\label{tab:data_aware}

\begin{tabular}{clcccccc}
\toprule
\multirow{2}{*}{\textbf{Antibiotic}} & \multirow{2}{*}{\textbf{Split}} 
& \multicolumn{2}{c}{\textbf{Total}} & \multicolumn{2}{c}{\textbf{R Label}} & \multicolumn{2}{c}{\textbf{\% R}} \\ 
& & Images & Cells & Images & Cells & Images & Cells \\
\midrule

\multirow{3}{*}{AZI} & Train & 82 & 22,218 & 59 & 15,610 & 72\% & 70\% \\
                     & Val   & 29 & 8,785  & 21 & 5,918  & 72\% & 67\% \\
                     & Test  & 27 & 7,460  & 20 & 5,538  & 74\% & 74\% \\ \midrule

\multirow{3}{*}{TE}  & Train & 77 & 29,265 & 51 & 18,480 & 66\% & 63\% \\
                     & Val   & 21  & 3,459  & 13  & 1,789  & 62\% & 52\% \\
                     & Test  & 25 & 11,991 & 17 & 9,396  & 68\% & 78\% \\ \midrule

\multirow{3}{*}{CIP} & Train & 88 & 23,123 & 50 & 16,565 & 57\% & 72\% \\
                     & Val   & 26 & 3,315  & 13 & 2,108  & 50\% & 64\% \\
                     & Test  & 28 & 8,470  & 20 & 5,453  & 71\% & 64\% \\ \bottomrule
\end{tabular}
\end{table}

Under both strategies, the class distribution across splits for all three antibiotics is moderately imbalanced, with resistant images accounting for approximately 50\% to 76\% of each split, reflecting the clinical prevalence of resistance among the collected isolates.

\section{Methodology}\label{sec:method}
% mô tả tổng thể toàn bộ pipeline, hình vẽ toàn bộ pipeline 
%\subsection{Overall Framework}
%\label{sec:overall}

\begin{figure}[h]
    \centering
    \includegraphics[width=\linewidth]{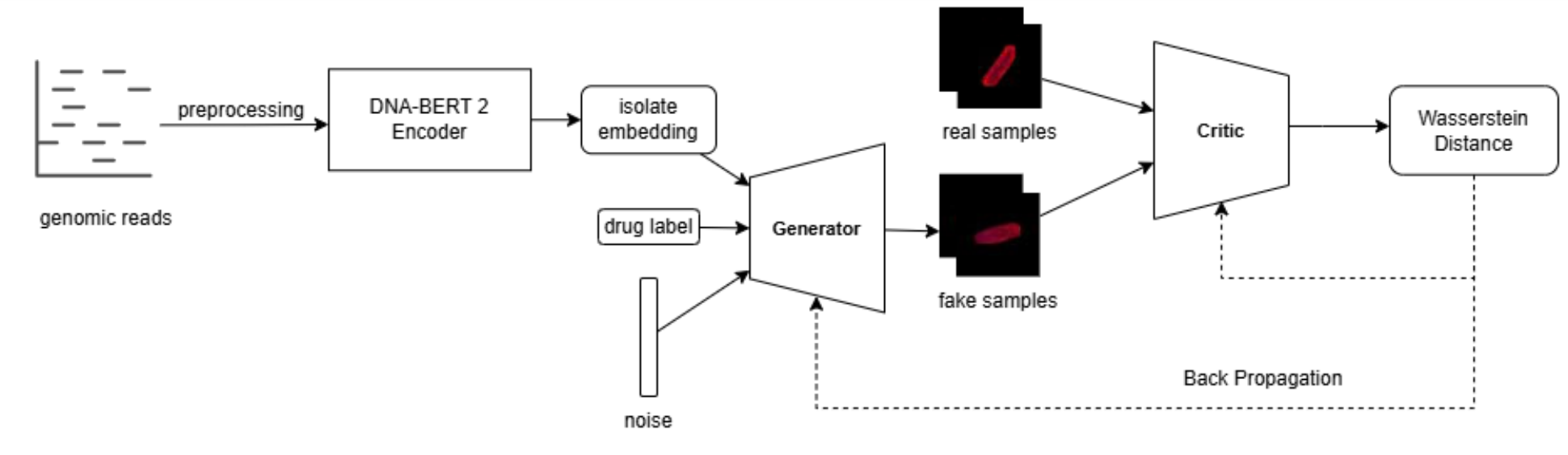}
    \caption{Overview of the proposed cWGAN-GP framework}
    \label{fig:pipeline}
\end{figure}
We propose a conditional generative framework that synthesizes colony
microscopy images of bacterial isolates from genomic and phenotypic
information.
Figure~\ref{fig:pipeline} provides an overview of the complete system.

The framework comprises two components.
The first is a \textbf{genomic encoding pipeline} that transforms raw
paired-end sequencing reads into a fixed-length 768-dimensional
isolate embedding using DNABERT-2, a transformer pretrained on
large-scale genomic corpora.
This embedding serves as the primary conditioning signal throughout
the generative model, encoding the genomic identity of the isolate
in a form that the GAN can consume directly.
The second is a conditional Wasserstein GAN with Gradient Penalty (cWGAN-GP) which generates synthetic colony images from the
isolate embedding, an antibiotic treatment label, and a sampled
latent noise vector, described in detail in
Section~\ref{sec:gan}.
The GAN is conditioned on all available biological context so that
the generated images reflect the combined effect of genomic
background, drug exposure, and resistance phenotype.

The GAN framework is developed across two training phases, each
serving a distinct purpose.
\textbf{Phase~1} is a backbone ablation study: seven pretrained CNN
architectures are evaluated as the visual feature extractor inside
the generator under identical training conditions, with the model
conditioned on the isolate embedding and the antibiotic label.
The best-performing backbone is selected based on FID, KID, and SSIM
computed on the test set.
\textbf{Phase~2} builds on Phase 1 by training the selected model (ConvNeXt-Tiny backbone~\cite{liu2022convnet}) with an additional condition: the antimicrobial resistance status of the isolate.
This extension enables the model to generate images that distinguish
between resistant and sensitive phenotypes under each antibiotic,
directly supporting the downstream resistance classification task
described in Section~\ref{sec:performance}.

\subsection{Genomic Sequence Encoding}
\label{sec:genomic}

\begin{figure}[h]
    \centering
    \includegraphics[width=\linewidth]{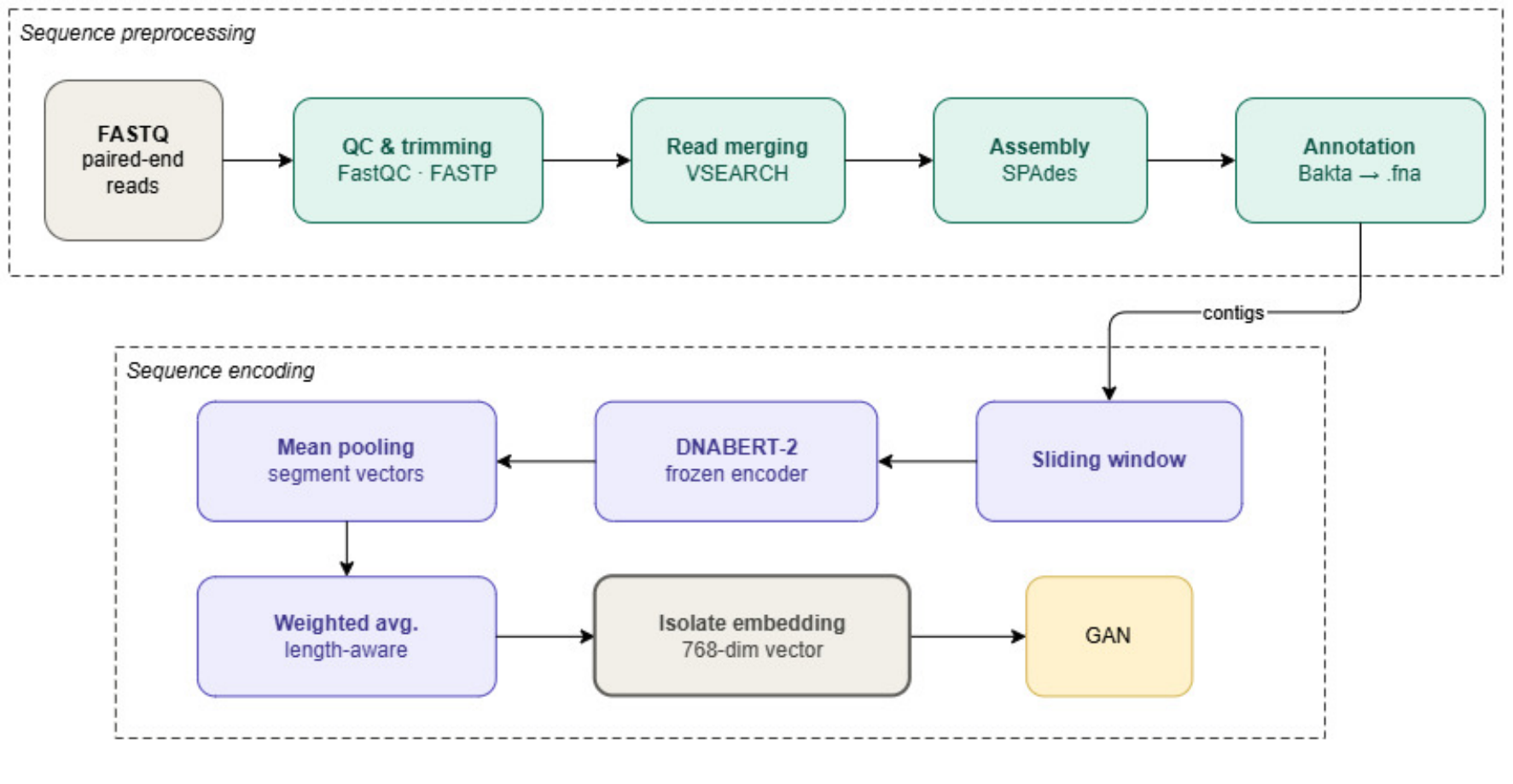}
    \caption{Genomic encoding pipeline.}
    \label{fig:genomic_encoding_pipeline}
\end{figure}

% mô tả bước xử lý gen để encode (thêm hình)
The complete genomic encoding pipeline is illustrated in
Figure~\ref{fig:genomic_encoding_pipeline}. It consists of two stages: a
sequence preprocessing stage that converts raw sequencing reads into clean,
annotated contigs, and a sequence encoding stage that maps those contigs into a
single fixed-length vector per isolate.

\paragraph{Sequence Preprocessing}
%\subsubsection{Sequence Preprocessing}
\begin{comment}
Raw paired-end sequencing reads in FASTQ format are processed through a
five-stage pipeline before being passed to the sequence encoder.
Each pair of raw FASTQ files is first evaluated using
FastQC~\cite{andrews2017fastqc} to inspect per-base quality scores, GC content distribution, sequence duplication levels, and potential adapter contamination.
Low-quality reads are subsequently filtered with FASTP~\cite{chen2018fastp}, where adapter sequences are detected automatically in paired-end mode and bases with a Phred quality score below 20 (estimated error probability $> 1\%$) are discarded.

Cleaned paired-end reads are then merged into single contiguous reads using VSEARCH~\cite{rognes2016vsearch}, requiring a minimum overlap of 168~bp between the forward and reverse strands with at most 5 mismatches permitted in the overlap region; merged reads shorter than 168~bp are discarded to eliminate spuriously short products. The merged reads for each isolate are subsequently assembled into genomic contigs using SPAdes~\cite{bankevich2012spades} in careful mode, which activates a post-assembly mismatch correction step.
Finally, assembled contigs are annotated using Bakta~\cite{schwengers2021bakta} with its lightweight database to produce sequences in \texttt{.fna} format, standardizing contig identifiers and sequence representation across all 22 isolates for consistent input to the encoding stage.
\end{comment}

Raw paired-end sequencing reads in FASTQ format are processed through a
five-stage preprocessing pipeline whose goal is to remove low-quality data and
reconstruct, for each isolate, a clean set of annotated genomic contigs. Each
pair of FASTQ files is first inspected with FastQC~\cite{andrews2017fastqc},
which reports per-base quality scores, GC content distribution, sequence
duplication levels, and potential adapter contamination. Based on this
assessment, the reads are cleaned with FASTP~\cite{chen2018fastp}: adapter
sequences are detected automatically in paired-end mode, and bases with a Phred
quality score below~20 (an estimated error probability above~$1\%$) are trimmed.
The cleaned forward and reverse reads of each pair are then merged into single
contiguous reads using VSEARCH~\cite{rognes2016vsearch}. Rather than employing a fixed threshold, the minimum overlap length required for the merging process is dynamically determined for each isolate based on the insert size distribution mean extracted from the FASTP post-filtering metadata. The algorithm allows at most five mismatches within this overlap region, and merged reads falling below the calculated adaptive threshold are discarded to eliminate spuriously short products. For each isolate, the merged reads are
assembled into genomic contigs using SPAdes~\cite{bankevich2012spades} in
careful mode, which enables an additional post-assembly step that
corrects mismatches and short indels. Finally, the assembled contigs are
annotated with Bakta~\cite{schwengers2021bakta} using its lightweight database,
producing sequences in \texttt{.fna} format and standardizing contig identifiers
and sequence representation across all 22~isolates for consistent input to the
encoding stage.

\paragraph{Sequence Encoding with DNABERT-2}

To represent each isolate as a dense vector, we use
DNABERT-2~\cite{zhou2024dnabert} as a frozen feature extractor, which lets us
reuse its pretrained genomic representations while avoiding the computational
cost of fine-tuning. Encoding proceeds through a hierarchy of three levels:
segment, contig, and isolate. Because transformer encoders accept only a limited
input length, each contig is first split into overlapping segments of 512~bp
using a sliding window with a stride of 256~bp, where the 256~bp overlap between
adjacent windows preserves contextual continuity across segment boundaries while
keeping each input within the encoder's length limit. Every segment is passed
independently through DNABERT-2 to obtain token-level contextual embeddings,
which are combined by mean pooling over the sequence-length dimension to yield
one fixed-length vector per segment. The segment vectors originating from the
same contig are then averaged to form a single contig-level representation
$E_{\text{contig}, i}$, and the contig-level vectors are in turn aggregated into
one isolate-level vector through a length-aware weighted average in which longer
contigs, being more information-rich, contribute proportionally more to the final
representation. The isolate embedding $E_{\text{isolate}}$ is computed as:
\begin{equation}
E_{\text{isolate}} = \sum_{i=1}^{n}  \frac{L_i}{\sum_{j=1}^{n} L_j} \cdot E_{\text{contig}, i} 
\end{equation}
where $n$ is the number of contigs in the isolate, $L_i$ is the length of the
$i$-th contig, and $E_{\text{contig}, i}$ is its representation. The result is a
768-dimensional vector that summarizes the genomic characteristics of the
isolate and serves as the conditioning input to the GAN framework.

\subsection{GAN-based Genome-to-Image Translation}
% mô tả các  GAN models đang dùng (thêm hình)
\label{sec:gan}
%\subsubsection{Problem Formulation}
\label{sec:formulation}
Let $\mathbf{x} \in [-1,1]^{3 \times 64 \times 64}$ denote a colony
microscopy image of a bacterial isolate, and let $\mathbf{c}$ denote
the set of biological conditioning signals associated with that image.
The objective of the generative model is to learn a generator $G$
and a critic $D$ such that the model distribution
$p_G(\mathbf{x} \mid \mathbf{c})$ approximates the true data
distribution $p_{\text{data}}(\mathbf{x} \mid \mathbf{c})$.
 
The conditioning context $\mathbf{c}$ differs between the two
training phases.
 
In \textbf{Phase~1}, $\mathbf{c} = (E, y)$ consists of two signals:
the \emph{isolate embedding} $E \in \mathbb{R}^{768}$ derived from
DNABERT-2 (Section~\ref{sec:genomic}), which encodes the genomic
identity of the bacterial strain; and the \emph{antibiotic treatment
label} $y \in \{0,\ldots,3\}$, corresponding to the four treatment
conditions in the dataset: no drug, azithromycin (AZI), tetracycline
(TE), and ciprofloxacin (CIP).
All four conditions are included to expose the model to the full
range of colony morphology across treatment states.
The Phase~1 objective is to learn $p_G(\mathbf{x} \mid E, y)$.
 
In \textbf{Phase~2}, $\mathbf{c} = (E, y, r)$ extends the Phase~1
conditioning with a third signal: the \emph{antimicrobial resistance
status} $r \in \{0, 1\}$, where $r=0$ denotes susceptibility and
$r=1$ denotes resistance of the isolate to the applied drug.
Since resistance status is clinically meaningful only in the presence
of a specific antibiotic, images collected under the no-drug
condition are excluded, reducing the drug label space to
$y \in \{0, 1, 2\}$ (AZI, TE, CIP).
The Phase~2 objective is to learn $p_G(\mathbf{x} \mid E, y, r)$.
\subsubsection{Generator Architecture}
\label{sec:generator}
 \begin{figure}[h]
    \centering
    \includegraphics[width=\linewidth]{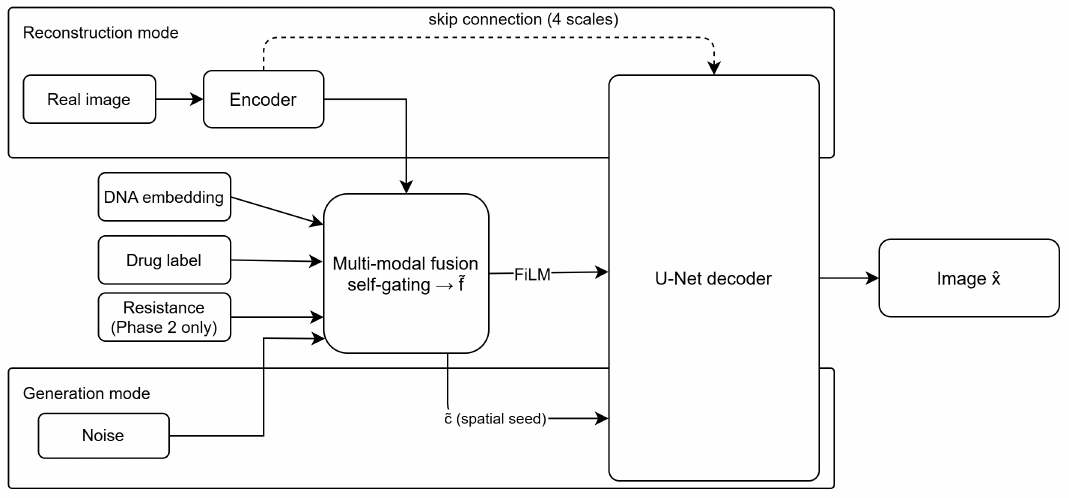}
    \caption{Overview of the Generator architecture}
    \label{fig:generator}
\end{figure}
Figure~\ref{fig:generator} illustrates the overall generator architecture, showing the three functional components: the multi-modal conditioning fusion module, the pretrained CNN backbone encoder, and the FiLM-conditioned U-Net decoder.
The generator operates in two modes during training: \emph{pure
generation} mode synthesizes images from noise $\mathbf{z}$ without
skip connections and is used for the adversarial objective;
\emph{reconstruction} mode encodes a real image through the backbone
to extract skip features before decoding, and is used for the
reconstruction objective.
 
\paragraph{Multi-modal conditioning fusion.}
In Phase~1, three independent branches encode the input signals into
a shared conditioning representation.
The \emph{noise branch} projects
$\mathbf{z} \sim \mathcal{N}(\mathbf{0}, \mathbf{I}_{100})$
through a two-layer MLP with ReLU, LayerNorm, and Dropout ($p{=}0.2$)
to $\mathbf{f}_z \in \mathbb{R}^{512}$.
The \emph{DNA branch} maps $E$ to
$\mathbf{f}_E \in \mathbb{R}^{512}$ via a three-layer MLP with ReLU,
LayerNorm, and Dropout ($p{=}0.2$).
The \emph{antibiotic branch} looks up a learnable embedding
$\mathbf{e}_y \in \mathbb{R}^{128}$ for label $y$ and projects it
to $\mathbf{f}_y \in \mathbb{R}^{256}$ through a two-layer MLP with
ReLU and LayerNorm.
The three outputs are concatenated to form
$\mathbf{f} = [\,\mathbf{f}_z \;\|\; \mathbf{f}_E \;\|\;
\mathbf{f}_y\,] \in \mathbb{R}^{1280}$.
 
In Phase~2, a fourth \emph{resistance branch} is added to the fusion
module: a learnable embedding $\mathbf{e}_r \in \mathbb{R}^{64}$ for
the resistance label $r$ is projected to
$\mathbf{f}_r \in \mathbb{R}^{128}$ through a two-layer MLP with
ReLU and LayerNorm, and concatenated with the other branches,
extending the fused vector to:
\begin{equation}
  \mathbf{f} = [\,\mathbf{f}_z \;\|\; \mathbf{f}_E \;\|\;
    \mathbf{f}_y \;\|\; \mathbf{f}_r\,] \in \mathbb{R}^{1408}.
  \label{eq:fusion_concat}
\end{equation}
In both phases, a self-gating mechanism scales $\mathbf{f}$
element-wise by a learned attention mask:
\begin{equation}
  \tilde{\mathbf{f}} = \mathbf{f} \odot
    \sigma\!\left(
      \mathbf{W}_2\,\tanh\!\left(\mathbf{W}_1\,\mathbf{f} +\mathbf{b}_1\right) + \mathbf{b}_2
    \right),
  \label{eq:gate}
\end{equation}
where $\mathbf{W}_1, \mathbf{W}_2 \in \mathbb{R}^{d \times d}$ are learned weight matrices with corresponding bias vectors $\mathbf{b}_1, \mathbf{b}_2 \in \mathbb{R}^d, d = 1280$ in Phase~1 and $d = 1408$ in Phase~2, and $\sigma$ is the sigmoid function.
The gated representation $\tilde{\mathbf{f}}$ is then projected to
$\tilde{\mathbf{c}} \in \mathbb{R}^{2048}$ through a two-layer MLP
with ReLU, LayerNorm, and Dropout ($p{=}0.2$).Within the decoder, $\tilde{\mathbf{c}}$ serves exclusively as the 
source of the spatial seed $\mathbf{s}$ in pure generation mode, 
while $\tilde{\mathbf{f}}$ provides the conditioning signal for 
all FiLM layers.
\paragraph{Pretrained CNN backbone encoder.}
The generator employs a pretrained CNN backbone to extract
hierarchical multi-scale visual features from the input image.
The backbone is modular and interchangeable; in Phase~1, seven
pretrained architectures are evaluated under identical conditions:
VGG-16~\cite{simonyan2014vgg},
ResNet-101~\cite{he2016resnet},
EfficientNet-B0~\cite{tan2019efficientnet},
DenseNet-121~\cite{huang2017densenet},
InceptionNet-V1~\cite{szegedy2015googlenet},
ConvNeXt-Tiny~\cite{liu2022convnet}, and
MobileNetV2~\cite{sandler2018mobilenetv2},
all initialised with ImageNet-pretrained weights.
The selection criterion and results of this ablation are reported
in Section~\ref{sec:results}.
In Phase~2, the backbone is fixed to ConvNeXt-Tiny, the
best-performing variant identified in Phase~1.
 
Regardless of backbone choice, feature maps are extracted at four
spatial downsampling scales.
For the ConvNeXt-Tiny backbone, the output channels are
$[96, 192, 384, 768]$ at spatial resolutions $16{\times}16$,
$8{\times}8$, $4{\times}4$, and $2{\times}2$ respectively
(for a $64{\times}64$ input).
The deepest feature map is further processed by a $3{\times}3$
convolution to 512 channels, followed by BatchNorm, ReLU, a
residual block, and a self-attention layer~\cite{zhang2019self},
yielding the spatial seed
$\mathbf{s} \in \mathbb{R}^{512 \times 2 \times 2}$.
In pure generation mode, the spatial seed is obtained from
$\tilde{\mathbf{c}}$ via a learned linear projection:
$\mathbf{s} = \mathrm{reshape}(\mathbf{W}_s\,\tilde{\mathbf{c}})$,
$\mathbf{W}_s : \mathbb{R}^{2048} \to \mathbb{R}^{2048}$.
 
\paragraph{FiLM-conditioned U-Net decoder.}
The decoder upsamples $\mathbf{s}$ through five transposed
convolution blocks to a $64{\times}64$ feature map.
Each block contains a transposed convolution ($4{\times}4$ kernel,
stride 2, padding 1), BatchNorm, ReLU, and a residual block.
The block at the $8{\times}8$ scale additionally contains a
self-attention layer.
In reconstruction mode, skip connections from the backbone are
concatenated with decoder activations at matching scales and merged
via $1{\times}1$ pointwise convolution; bilinear interpolation is
applied when spatial dimensions do not match.
Conditioning is injected at the first four decoder stages 
(4×4, 8×8, 16×16, and 32×32) via Feature-wise Linear 
Modulation (FiLM)~\cite{perez2018film}:
\begin{equation}
  \tilde{\mathbf{h}} = (1 + \boldsymbol{\gamma}) \odot \mathbf{h}
    + \boldsymbol{\beta},
  \label{eq:film}
\end{equation}
where  $(\boldsymbol{\gamma}, \boldsymbol{\beta}) \in \mathbb{R}^{C}$ are the modulation parameters and $C$ is the number of channels at that scale. Both parameters are jointly predicted from $\tilde{\mathbf{f}}$ 
by a two-layer MLP with ReLU activation that projects $\tilde{\mathbf{f}}$ 
to $\mathbb{R}^{2C}$, whose output is split evenly into 
$\boldsymbol{\gamma}$ and $\boldsymbol{\beta}$.
The output layer applies two $3{\times}3$ convolutions
($32{\to}16{\to}3$ channels) with ReLU and Tanh, producing
$\hat{\mathbf{x}} \in [-1,1]^{3\times64\times64}$.

\subsubsection{Critic Architecture}
\label{sec:discriminator}

\begin{figure}[h]
    \centering
    \includegraphics[width=\linewidth]{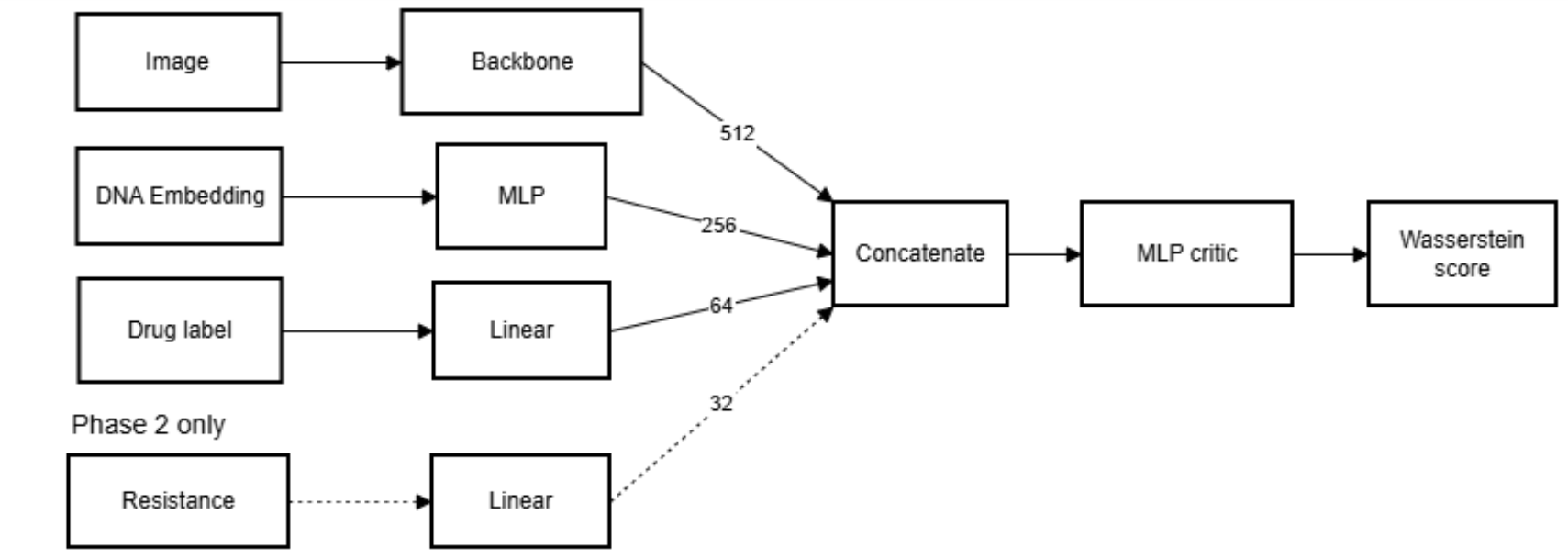}
    \caption{Overview of the Critic architecture}
    \label{fig:critic}
\end{figure}

The Critic $D$ is a multi-modal critic that jointly evaluates
an image against the full conditioning context $\mathbf{c}$, thereby penalising outputs that are either visually unrealistic or
inconsistent with their associated biological context.
The design rationale and the training instabilities that motivated
this choice are discussed in Section~\ref{sec:objectives}.
Figure~\ref{fig:critic} illustrates the critic architecture, 
in which image features and conditioning signals are processed 
in parallel branches before being concatenated and passed through 
a shared MLP to produce a single Wasserstein score.

The critic backbone processes the input image as a separate
instance from the generator backbone and returns the deepest feature
map, which is flattened and projected to $\mathbb{R}^{512}$ via a
linear layer with LeakyReLU ($\alpha{=}0.2$) and Dropout ($p{=}0.4$).
 
In Phase~1, two conditioning branches process the non-image inputs
in parallel alongside the image features:
the \emph{DNA branch} maps $E$ to $\mathbb{R}^{256}$ via a two-layer
MLP with LeakyReLU and Dropout ($p{=}0.3$);
and the \emph{antibiotic branch} encodes $y$ as a $K$-dimensional
one-hot vector and projects it to $\mathbb{R}^{64}$ with LeakyReLU
and Dropout ($p{=}0.2$).
The three representations are concatenated to
$832$ dimensions and passed through a three-layer MLP
critic head with LeakyReLU and Dropout ($p{=}0.5$) to produce a
single real-valued Wasserstein score without sigmoid activation.
 
In Phase~2, a third \emph{resistance branch} is added: the
resistance status $r$ is encoded as a two-dimensional one-hot vector
and projected to $\mathbb{R}^{32}$ with LeakyReLU and Dropout
($p{=}0.2$), extending the concatenated representation to
$864$ dimensions.
 
\paragraph{Backbone selection for the critic.}
Computing the gradient penalty under WGAN-GP requires extracting
per-sample gradients from interpolated inputs
(Section~\ref{sec:objectives}).
Batch Normalization (BN) computes mean and variance statistics
across the entire mini-batch, introducing inter-sample information
leakage that corrupts the independent per-sample gradient signal
required by the penalty term.
Any backbone containing BN layers is therefore mathematically
incompatible with gradient penalty computation and cannot be used
inside the critic.
 
Among the seven backbone architectures evaluated in Phase~1, Batch Normalization layers are present in ResNet-101, EfficientNet-B0, DenseNet-121, InceptionNet-V1, and MobileNetV2. For these five cases, the critic backbone is substituted with ConvNeXt-Tiny, which replaces Batch Normalization with Layer Normalization and is therefore compatible with per-sample gradient computation.
For the remaining two backbones (ConvNeXt-Tiny and VGG-16) the critic uses the same backbone as the generator, as neither architecture contains Batch Normalization. This substitution policy ensures gradient penalty validity across all ablation experiments while keeping the critic capacity as consistent as possible with the generator.
\subsubsection{Training Objectives}
\label{sec:objectives}

Both phases adopt the Wasserstein GAN with gradient penalty
(WGAN-GP)~\cite{gulrajani2017improved} as the adversarial training
framework.
WGAN-GP optimizes the Wasserstein-1 distance between the real and
generated distributions, enforcing a Lipschitz constraint on the
critic via a gradient penalty on interpolated samples rather than
weight clipping.
We select WGAN-GP over the original GAN
formulation~\cite{goodfellow2014gan} for three reasons specific to
our setting.
First, training stability is critical when the conditioning space
is high-dimensional: the 768-dimensional genomic embedding combined
with discrete treatment and resistance labels creates a complex
conditional distribution that is prone to mode collapse and vanishing
gradients under a standard GAN critic.
Second, the Wasserstein distance provides a smoother and more
informative gradient signal than the Jensen--Shannon divergence,
which is particularly beneficial in the early stages of training
when the generator has not yet learned to produce recognizable
colony structure and the real and generated distributions have
limited overlap.
Finally, the gradient penalty enforces the Lipschitz constraint
without restricting the critic architecture, allowing us to use a pretrained ConvNeXt-Tiny backbone inside the critic without
weight clipping interference.
 
\paragraph{Training instabilities and iterative solutions.}
Two failure modes were encountered during preliminary training and
resolved through successive interventions, motivating the final
objective design.
 
In the \emph{first failure mode}, training with only the adversarial
objective caused the generator to consistently produce entirely black
images — a degenerate solution that satisfies the Wasserstein
distance in early training before the critic has learned to provide
informative gradients.
A pixel-level reconstruction loss was introduced to resolve this,
anchoring the generator output to real colony appearance from the
first training epoch.
 
In the \emph{second failure mode}, adding the reconstruction loss
eliminated the all-black output but introduced conditioning-space
mode collapse: the generator converged to a single ``average colony''
image regardless of isolate identity or antibiotic condition,
satisfying reconstruction error while ignoring all conditioning
signals.
This was resolved by adopting a \emph{conditional} WGAN-GP
formulation in which the critic receives the full conditioning
context $\mathbf{c}$ alongside the image, forcing the critic to
jointly evaluate realism and conditioning consistency and thereby
driving inter-class diversity.
 
\paragraph{Adversarial loss.}
Training follows the WGAN-GP protocol~\cite{gulrajani2017improved}
with $n_{\text{critic}} = 5$ critic updates per generator
step.
The critic objective is:
\begin{equation}
  \mathcal{L}_D =
    -\mathbb{E}_{\mathbf{x} \sim p_{\text{data}(x|c)}}
      \bigl[D(\mathbf{x},\, \mathbf{c})\bigr]
    + \mathbb{E}_{\mathbf{z} \sim p_z,c}
      \bigl[D(G(\mathbf{z},\, \mathbf{c}),\, \mathbf{c})\bigr]
    + \lambda_{\text{GP}}\,\mathcal{L}_{\text{GP}},
  \label{eq:critic_loss}
\end{equation}
where the gradient penalty is computed on interpolated samples
$\hat{\mathbf{x}} = \alpha\,\mathbf{x} +
(1{-}\alpha)\,G(\mathbf{z}, \mathbf{c})$,
$\alpha \sim \mathcal{U}[0,1]$:
\begin{equation}
  \mathcal{L}_{\text{GP}} =
    \mathbb{E}_{{\hat{\mathbf{x}},c}}\!\left[
      \Bigl(
        \bigl\|\nabla_{\hat{\mathbf{x}}}
          D(\hat{\mathbf{x}},\, \mathbf{c})
        \bigr\|_2 - 1
      \Bigr)^{\!2}
    \right],
  \label{eq:gp}
\end{equation}
with $\lambda_{\text{GP}} = 10$.
 
\paragraph{Reconstruction loss.}
For each real training image $\mathbf{x}$, the generator backbone
encodes $\mathbf{x}$ to extract multi-scale skip features, and the
decoder reconstructs $\mathbf{x}$ under the correct conditioning:
\begin{equation}
  \mathcal{L}_{\text{recon}} =
    \bigl\|G_{\text{recon}}(\mathbf{x},\, \mathbf{c})
    - \mathbf{x}\bigr\|_1,
  \label{eq:recon}
\end{equation}
where $G_{\text{recon}}$ denotes the encode-then-decode path.
 
\paragraph{Total generator objective.}
\begin{equation}
  \mathcal{L}_G =
    -\mathbb{E}_{\mathbf{z}}\!
      \bigl[D(G(\mathbf{z},\, \mathbf{c}),\, \mathbf{c})\bigr]
    + \lambda_{\text{recon}}\,\mathcal{L}_{\text{recon}},
  \label{eq:g_total}
\end{equation}
with $\lambda_{\text{recon}} = 100$.
The high reconstruction weight ensures pixel-level stability during
early training while the conditional critic drives inter-class
diversity as training progresses.
For each training step, the critic is updated $n_{\text{critic}}=5$ times using real samples and generated samples from the pure generation path. The generator is then updated once using a combined objective consisting of adversarial loss from the pure generation path and reconstruction loss from the reconstruction path. The reconstruction path is used only during training to stabilize optimization and is disabled at validation and test time, where only the pure generation path is used for image synthesis.

\subsubsection{Implementation Details}
\label{sec:impl}

Both the generator and critic are optimized using
Adam~\cite{kingma2015adam} with $\beta_1 = 0$, $\beta_2 = 0.9$,
and learning rate $10^{-4}$ for both networks in both phases.
Training runs for a maximum of 50 epochs with batch size 512,
latent dimension $d_z = 100$, and image resolution $64{\times}64$.

\paragraph{Image preprocessing.}
Prior to model input, each image undergoes adaptive contrast
enhancement.
The mean pixel intensity $\bar{I}$ is computed over the grayscale
channel and a contrast scale factor is derived as:
\begin{equation}
  \alpha = \frac{5}{4\,\max(\bar{I},\,\epsilon) + \epsilon} + 1,
  \label{eq:contrast}
\end{equation}
where $\epsilon = 10^{-2}$ prevents division by zero.
Darker images (lower $\bar{I}$) receive stronger enhancement;
brighter images are adjusted more conservatively.
The enhanced image is converted to RGB and normalised to $[-1, 1]$
per channel with mean $(0.5, 0.5, 0.5)$ and standard deviation
$(0.5, 0.5, 0.5)$.

\paragraph{Checkpoint selection in Phase~1.}
During the backbone ablation study, the best checkpoint for each
backbone variant is selected using the \emph{mean per-class FID}
on the validation set.
FID is evaluated independently for each (isolate, antibiotic) class
pair at the end of every epoch; the checkpoint achieving the lowest
average across all valid classes (those with at least 10 validation
samples) is retained as the representative of that backbone.
Per-class averaging is preferred over a global FID because the
dataset is class-imbalanced: a global metric is dominated by
high-frequency classes and fails to penalise poor performance on
rarer isolate--drug combinations.

\paragraph{Checkpoint selection in Phase~2.}
The final model (ConvNeXt-Tiny backbone with resistance
conditioning) employs a \emph{min-max FID} checkpoint selection
strategy.
At each epoch, the maximum per-class FID across all valid classes
is computed:
\begin{equation}
  \text{FID}_{\max} = \max_{k \in \mathcal{C}}\,\text{FID}_k,
  \label{eq:max_fid}
\end{equation}
and a checkpoint is saved whenever $\text{FID}_{\max}$ improves.
This strategy focuses training on the worst-performing class at
each epoch rather than on the class average, ensuring that no
(isolate, antibiotic) combination is neglected - a property
essential for the downstream resistance classification task, which
must generalise across all class combinations.
Early stopping monitors
$\text{FID}_{\max}$ as the tracked metric.

\section{Experimental Results}\label{sec:experiments}
% \subsection{Quantitative Evaluation Metrics}
% \label{sec:metrics}
Image quality is evaluated by two widely adopted metrics for generative models: the Fréchet Inception Distance (FID), introduced by Heusel et al. in 2017 \cite{heusel2017gans}, and the Kernel Inception Distance (KID), proposed by Bińkowski et al. in 2018 \cite{binkowski2018demystifying}. 
Both metrics are computed from feature representations extracted from the final average pooling layer of a pre-trained Inception-v3 network.

FID assumes that the extracted features follow multivariate Gaussian distributions and computes the Fréchet distance between the real and generated feature distributions using their means and covariance matrices:

\[\text{FID}(R,G) = |\mu_R - \mu_G|^2_2 + \text{Tr}\left(\Sigma_R + \Sigma_G - 2\sqrt{\Sigma_R \Sigma_G}\right)\]

where $\mu_R$ and $\mu_G$ denote the feature means, and $\Sigma_R$ and $\Sigma_G$ denote the corresponding covariance matrices of the real and generated distributions.

In contrast, KID does not rely on Gaussian assumptions and instead computes the squared Maximum Mean Discrepancy (MMD) between the two feature distributions:

\[\text{KID}(R, G) = MMD^2 = \mathbb{E}_{x,x' \sim R} [k(x, x')] + \mathbb{E}_{y,y' \sim G} [k(y, y')] - 2\mathbb{E}_{x \sim R, y \sim G} [k(x, y)]\]

where $\mathbb{E}$ denotes expectation and $k(x,y) = (\frac{1}{d}x^\top y + 1)^3$ is the polynomial kernel function, with $d$ representing the feature dimension.

Unlike FID's reliance on mean and covariance, KID is sensitive to higher-order distributional differences through its kernel formulation and provides an unbiased estimator even for relatively small sample sizes.
Lower FID/KID scores indicate better similarity between synthetic and real distributions. 

Additionally, we report the Structural Similarity Index Measure (SSIM) \cite{wang2004ssim} to assess the perceptual similarity between generated and real bacterial cells. Unlike FID and KID, which evaluate distribution-level similarity, SSIM measures image-level similarity based on luminance, contrast, and structural information:

\(\text{SSIM}(x,y) = [l(x,y)]^\alpha \cdot [c(x,y)]^\beta \cdot [s(x,y)]^\gamma\)
where: \\
\[
l(x,y) = \frac{2\mu_x\mu_y + C_1}{\mu_x^2 + \mu_y^2 + C_1}, \quad
c(x,y) = \frac{2\sigma_x\sigma_y + C_2}{\sigma_x^2 + \sigma_y^2 + C_2}, \quad
s(x,y) = \frac{\sigma_{xy} + C_3}{\sigma_x\sigma_y + C_3}
\]

For evaluation, each generated image is compared with all real images belonging to the same isolate-antibiotic class. The SSIM score of a generated image is defined as the average SSIM across these pairwise comparisons. The final SSIM reported for a model is obtained by averaging over all generated images.
Higher SSIM values indicate that generated cells exhibit luminance, contrast, and structural characteristics similar to those observed in real bacterial cells. However, because SSIM measures pairwise similarity between images, it does not directly assess distributional diversity and is therefore reported alongside FID and KID.

% \subsection{Backbone Ablation Results}
% \label{sec:results}

\subsection{Backbone Comparison Results}
\label{sec:results}
\begin{table*}[htbp]
\centering
\caption{Quantitative comparison of different backbones within cWGAN-GP framework on the test set. For the FID and KID, lower is better. For the SSIM, higher is better.}
\label{tab:metrics}
\renewcommand{\arraystretch}{1.2}
\begin{tabular}{c c c c}
\hline
\textbf{Backbone type} & \textbf{FID} & \textbf{KID} & \textbf{SSIM Avg} \\
\hline

ConvNeXt-Tiny   & \textbf{80.88} & \textbf{0.026} & 0.803 $\pm$ 0.058 \\
DenseNet121     & 82.18 & 0.027 & 0.808 $\pm$ 0.052 \\
MobileNet-V2    & 84.73 & 0.029 & 0.805 $\pm$ 0.056 \\
ResNet101       & 89.47 & 0.035   & 0.787 $\pm$ 0.062 \\
EfficientNet-B0 & 90.42 & 0.028 & \textbf{0.809 $\pm$ 0.055} \\
VGG-16          & 102.79 & 0.030 & 0.782 $\pm$ 0.071 \\
InceptionNet-V1 & 233.64 & 0.159 & \textbf{0.809 $\pm$ 0.064} \\
\hline
\end{tabular}
\end{table*}
The performance of the proposed cWGAN-GP architecture with seven different pretrained CNN backbones is assessed by the evaluation metrics described above. These backbones serve as the visual feature extractor within the generator, providing the multi-scale image representations from which the decoder synthesizes colony images; the critic backbone follows the substitution policy described in Section~\ref{sec:discriminator}. The evaluated architectures include ConvNeXt-Tiny, DenseNet121, MobileNet-V2, ResNet101, EfficientNet-B0, VGG-16, and InceptionNet-V1.

The quantitative results are summarized in Table~\ref{tab:metrics}. 
According to the distribution-based metrics, ConvNeXt-Tiny achieved the lowest FID (80.88) and KID (0.026) scores among all evaluated backbones, indicating the closest alignment between the generated and real image distributions. DenseNet121 and MobileNet-V2 produced similarly competitive results, whereas ResNet101 and EfficientNet-B0 exhibited moderately higher FID and KID values. In contrast, VGG-16 and InceptionNet-V1 yielded substantially worse scores, suggesting that these architectures were less effective as feature extractors within the proposed cWGAN-GP framework. Notably, FID and KID exhibited a consistent ranking across all backbones, indicating strong agreement between the two distribution-based evaluation metrics.

In terms of image-level similarity, EfficientNet-B0 and InceptionNet-V1 achieved the highest average SSIM scores (0.809), and the remaining backbones produced comparable SSIM values. 
SSIM remained relatively stable (0.782–0.809), suggesting that most generated images preserved similar local structural characteristics of real bacterial microscopy images, despite large differences in distributional realism (as FID and KID varied substantially across backbones).

However, the discrepancy between SSIM and the distribution-based metrics is particularly evident for EfficientNet-B0 and InceptionNet-V1, which achieved among the highest SSIM scores despite exhibiting substantially worse FID and KID values. This observation highlights an important limitation of SSIM: although it measures perceptual similarity between individual image pairs, it does not directly evaluate distributional realism or sample diversity. Consequently, SSIM is reported as a complementary reference metric, whereas FID and KID are considered the primary criteria for backbone selection.

\begin{figure}[htbp]
    \centering
    \begin{subfigure}[b]{0.33\textwidth}
        \centering
        \includegraphics[width=\linewidth,trim=0pt 4pt 0pt 0pt,clip]{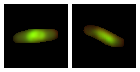}
        \includegraphics[width=\linewidth,trim=0pt 4pt 0pt 0pt,clip]{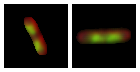}
        \includegraphics[width=\linewidth,trim=0pt 4pt 0pt 0pt,clip]{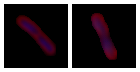}
        \includegraphics[width=\linewidth,trim=0pt 4pt 0pt 0pt,clip]{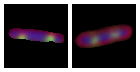}

        \caption{}
    \end{subfigure}%
    \begin{subfigure}[b]{0.33\textwidth}
        \centering
        \includegraphics[width=\linewidth,trim=0pt 4pt 0pt 0pt, clip]{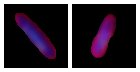}
        \includegraphics[width=\linewidth,trim=0pt 4pt 0pt 0pt, clip]{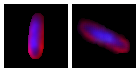}
        \includegraphics[width=\linewidth,trim=0pt 4pt 0pt 0pt, clip]{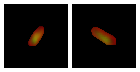}
        \includegraphics[width=\linewidth,trim=0pt 4pt 0pt 0pt, clip]{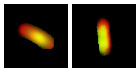}

        \caption{}
    \end{subfigure}%
    \begin{subfigure}[b]{0.33\textwidth}
        \centering
        \includegraphics[width=\linewidth,trim=0pt 4pt 0pt 0pt, clip]{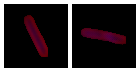}
        \includegraphics[width=\linewidth,trim=0pt 4pt 0pt 0pt, clip]{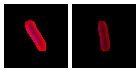}
        \includegraphics[width=\linewidth,trim=0pt 4pt 0pt 0pt, clip]{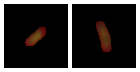}
        \includegraphics[width=\linewidth,trim=0pt 4pt 0pt 0pt, clip]{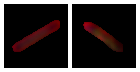}

        \caption{}
    \end{subfigure}%
    
    \caption{Examples of bacterial cells generated by the proposed cWGAN-GP with a ConvNeXt-Tiny backbone. Each synthetic image (right) was generated under the same genomic and phenotypic condition as the corresponding real image (left). Results are shown for (a) AZI, (b) TE, and (c) CIP.
}
    \label{fig:qualitative_samples}
\end{figure}

Based on the consistently superior FID and KID performance, ConvNeXt-Tiny was selected as the generator backbone for all subsequent experiments.
Figure~\ref{fig:qualitative_samples} shows representative real and synthetic cells produced by this configuration under each antibiotic.

% \subsection{Synthetic Image Quality}
% \label{sec:quality}
\subsection{Downstream Classification Performance}
\label{sec:performance}

To evaluate the practical utility of our generative model, we examine whether synthetic images from our cWGAN-GP can improve classification performance when real training data is scarce.
Three independent binary classification tasks are defined, one per antibiotic (AZI, TE, CIP), predicting resistant (R) vs.\ susceptible (S) phenotypes for each drug separately.
All classifiers use a fully fine-tuned ConvNeXt-Tiny backbone with a lightweight MLP head.
Each drug-specific classifier is tuned independently over class-weighted cross-entropy, dropout rate, backbone and head learning rates, and input noise intensity, with early stopping on the validation set.

% \paragraph{Data partitioning.}
All splits are performed at the microscopy-image level, so that cell images originating from the same microscopy image never appear in more than one split. We consider two protocols. 
(i)~\textit{Isolate-agnostic splitting}: images from all isolates are pooled and partitioned randomly, so the same isolate may appear in all three splits.
(ii)~\textit{Isolate-aware splitting}: 80\% of the 22 isolates are assigned to the training pool and the remaining isolates form the test set; the validation set is drawn from 25\% of the training-pool images, so a training isolate may contribute to both training and validation but never to testing. This protocol measures generalization to unseen isolates. 

% \paragraph{Low-data simulation and augmentation.}
% To simulate a low-data regime, we reduce the real training cells to 25\% per image - selected after a preliminary study at 100\%, 50\% and 25\% confirmed that performance degrades proportionally, providing sufficient headroom for augmentation to produce a measurable effect. 
% The number of images remains unchanged; only the cell count per image is reduced. This reduced dataset serves as the baseline training set.
To simulate a low-data regime, we reduce the real training cells to 25\% per image. This ratio was selected after a preliminary study at 100\%, 50\% and 25\% confirmed that performance degrades proportionally, leaving sufficient headroom for augmentation to produce a measurable effect.
The number of images remains unchanged; only the cell count per image is reduced. This reduced dataset serves as the baseline training set.
For augmentation, we train a cWGAN-GP variant extended with an additional conditioning branch for resistance/susceptibility labels.
Synthetic cells conditioned on antibiotic type and resistance status are added to the baseline training set. The number of synthetic cells is set to 50\% of the number of real cells retained in the baseline training set, forming the GAN-augmented training set.
Final evaluation for both conditions is performed exclusively on the held-out real test set, containing no synthetic samples, so that reported metrics reflect generalization to real microscopy images.

Classifier performance is evaluated using several standard metrics, including Precision, Recall, \mbox{F1-Score}, Accuracy, and AUC. 
These metrics provide complementary perspectives on classification performance by measuring prediction correctness, class sensitivity, and the balance between precision and recall.
The corresponding definitions are given below:

\begin{align}
\text{Accuracy}  &= \frac{\text{TP} + \text{TN}}{\text{TP} + \text{FP} + \text{TN} + \text{FN}} \\
\text{Precision} &= \frac{\text{TP}}{\text{TP} + \text{FP}} \\
\text{Recall} &= \frac{\text{TP}}{\text{TP} + \text{FN}} \\ 
\text{\mbox{F1-Score}}  &= 2 \times \frac{\text{Precision} \times \text{Recall}}
                  {\text{Precision} + \text{Recall}}
% \text{Macro Recall} &= \frac{\text{Recall}_R + \text{Recall}_S}{2}
\end{align}

% Although all metrics are reported, Recall remains particularly relevant in antimicrobial susceptibility testing because resistant isolates incorrectly classified as susceptible (false susceptibility) may lead to inappropriate treatment decisions.

The complete image generation, training, and evaluation pipeline was repeated using five different random seeds to assess performance. For each seed, the baseline and GAN-augmented models were trained and evaluated using identical data splits, enabling paired statistical comparisons.

\begin{figure}[htbp]
    \centering
    \begin{subfigure}[b]{0.32\textwidth}
        \centering
        \includegraphics[width=\linewidth, trim=1.5cm 2.5cm 2cm 1.5cm, clip]{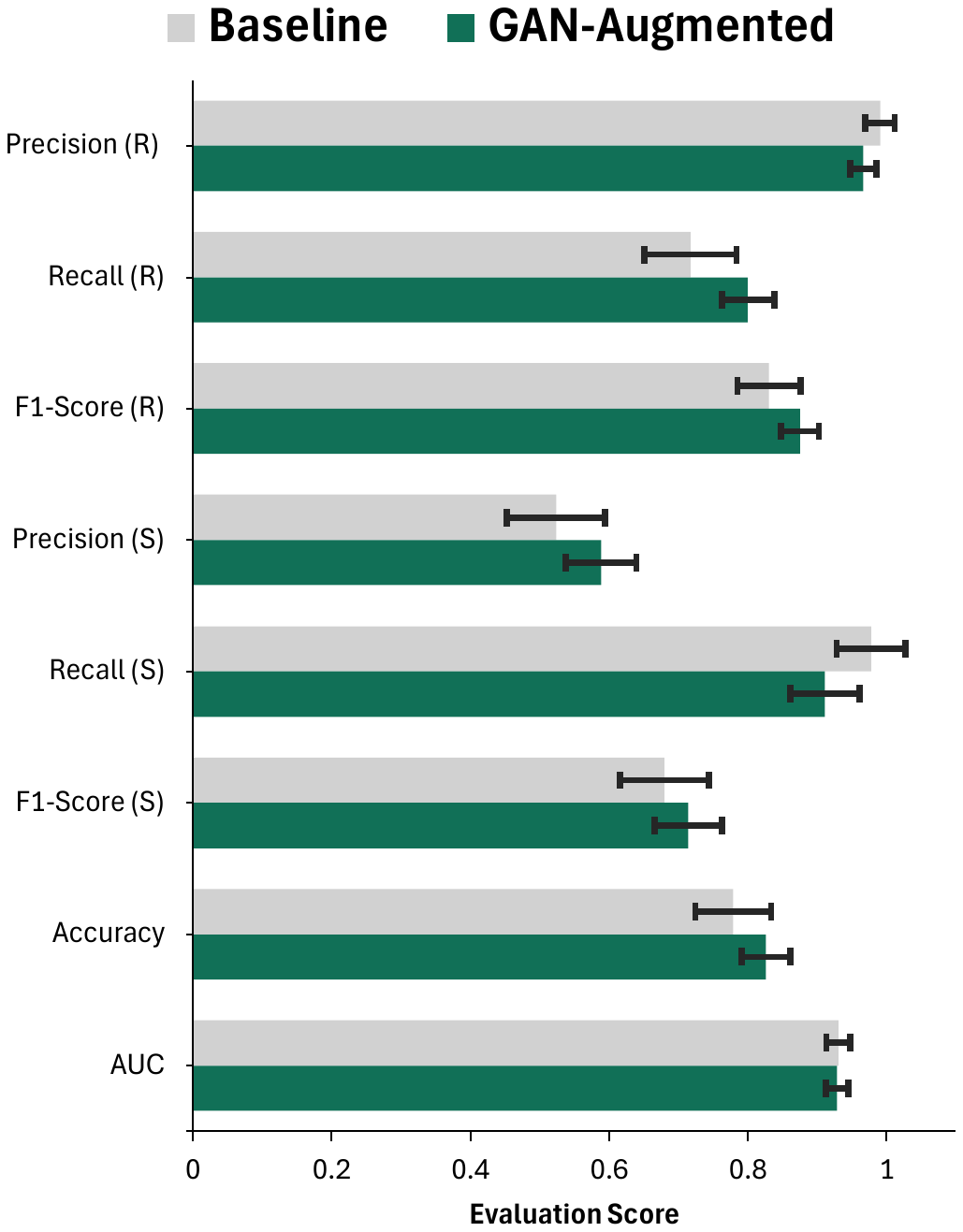}
        \caption{}
        \label{fig:chart_azi}
    \end{subfigure}
    \begin{subfigure}[b]{0.32\textwidth}
        \centering
        \includegraphics[width=\linewidth, trim=1.5cm 2.5cm 2cm 1.5cm, clip]{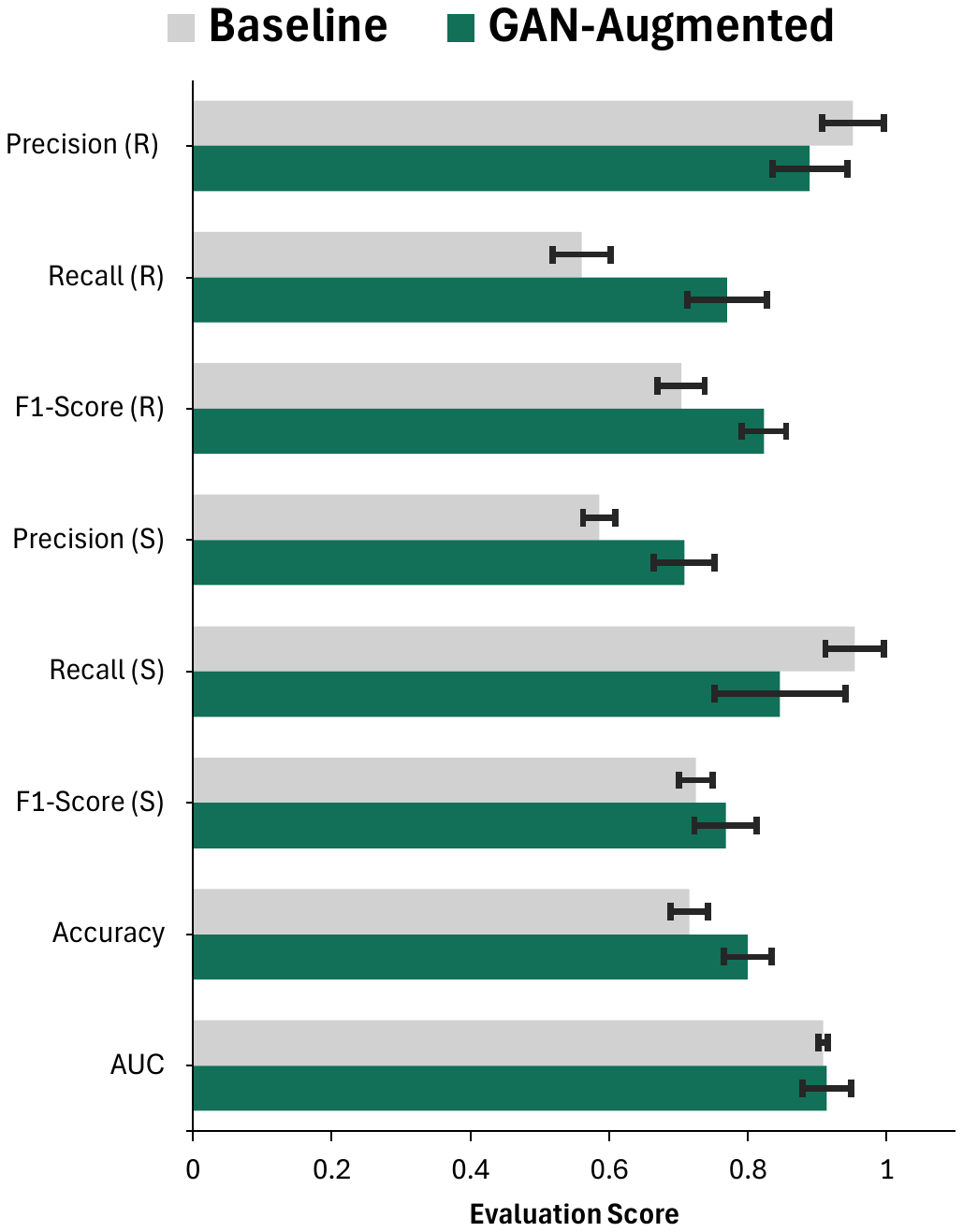}
        \caption{}
        \label{fig:chart_TE}
    \end{subfigure}
    \begin{subfigure}[b]{0.32\textwidth}
        \centering
        \includegraphics[width=\linewidth, trim=1.5cm 2.5cm 2cm 1.5cm, clip]{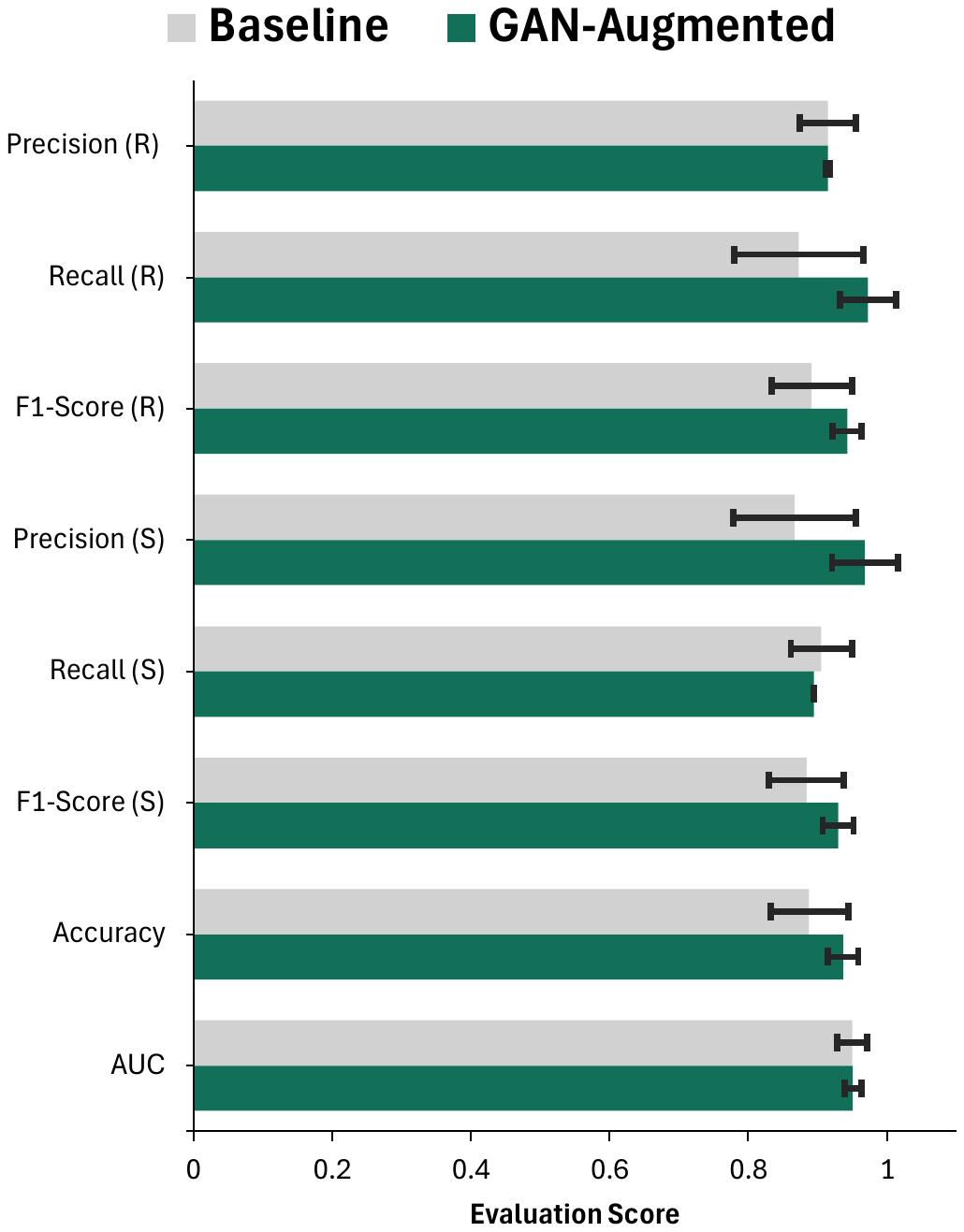}
        \caption{}
        \label{fig:chart_cip}
    \end{subfigure}
    \caption{(Isolate-Agnostic Splitting) Performance comparison between Baseline and GAN-Augmented training for (a) AZI, (b) TE, and (c) CIP.}
    \label{fig:chart_agnostic}
\end{figure}

% \subsubsection{Isolate-Agnostic Splitting}
Table~\ref{tab:metrics_random} and Figure~\ref{fig:chart_agnostic} report the results when isolates may be shared across splits.
Baseline performance is already high (AUC $\geq 0.9$ for all antibiotics) and stable across seeds.
GAN augmentation improved most mean scores, with significant gains in four metrics for AZI, four for TE, and two for CIP.
\mbox{Recall (R)} increased significantly for all three antibiotics, by $0.083$, $0.210$, and $0.100$ for AZI ($p=0.012$), TE ($p=0.0005$), and CIP ($p=0.031$), respectively.
Accuracy improved by $0.047$ for AZI ($p=0.044$) and $0.085$ for TE ($p=0.004$), and by $0.049$ for CIP, which was not significant ($p=0.065$). 
\mbox{F1-Score} (R) improved significantly for AZI and TE ($0.044$ and $0.119$), and \mbox{Precision (S)} for all three antibiotics ($0.065$, $0.123$ and $0.101$).

These gains come with a trade-off. 
\mbox{Precision (R)} decreased for AZI ($0.023$) and TE ($0.062$), and \mbox{Recall (S)} decreased for AZI ($0.067$) and TE ($0.108$), while AUC remained essentially unchanged (differences $\leq 0.006$; $p>0.38$ where tested).
This suggests that the augmented models operate at a different point on a similar ROC curve, with a decision boundary shifted toward higher sensitivity for the resistant class: resistance is flagged more readily, at the cost of occasionally misclassifying susceptible isolates.
In clinical practice this trade-off may be acceptable, as missing a resistant case (false susceptibility) generally has more severe consequences than a false resistance call.

\begin{table*}[htbp]
\centering
\caption{(Isolate-Agnostic Splitting) Per-antibiotic comparison between Baseline and GAN-Augmented training. Image-level hard voting, mean $\pm$ std over 5 seeds. \textbf{Bold}: numerically higher score within each metric for the same antibiotic. $* = p < 0.05$ (paired one-sided $t$-test, $H_1$: Baseline $<$ GAN-Augmented). }
\label{tab:metrics_random}

\renewcommand{\arraystretch}{1.15}
\begin{tabular}{c l c c c}
\toprule
\textbf{Antibiotic} & \textbf{Metric} & \textbf{Baseline} & \textbf{GAN-Augmented} & \textbf{$p$-value} \\
\midrule

\multirow{8}{*}{AZI} 
& \mbox{Precision (R)}  & 0.990 $\pm$ 0.021 & 0.967 $\pm$ 0.019          & -- \\
& \mbox{Recall (R)}     & 0.717 $\pm$ 0.066 & \textbf{0.800 $\pm$ 0.038} & 0.0120* \\
& \mbox{F1-Score} (R)   & 0.831 $\pm$ 0.046 & \textbf{0.875 $\pm$ 0.027} & 0.0304* \\
& \mbox{Precision (S)}  & 0.523 $\pm$ 0.071 & \textbf{0.588 $\pm$ 0.051} & 0.0389* \\
& \mbox{Recall (S)}     & 0.978 $\pm$ 0.050 & 0.911 $\pm$ 0.050          & -- \\
& \mbox{F1-Score} (S)   & 0.680 $\pm$ 0.064 & \textbf{0.714 $\pm$ 0.049} & 0.1410 \\
& Accuracy       & 0.779 $\pm$ 0.055 & \textbf{0.826 $\pm$ 0.035} & 0.0438* \\
& AUC            & 0.930 $\pm$ 0.017 & 0.929 $\pm$ 0.016          & -- \\
\midrule
 
\multirow{8}{*}{TE}
& \mbox{Precision (R)}  & 0.951 $\pm$ 0.045 & 0.889 $\pm$ 0.054          & -- \\
& \mbox{Recall (R)}     & 0.560 $\pm$ 0.042 & \textbf{0.770 $\pm$ 0.057} & 0.0005* \\
& \mbox{F1-Score} (R)   & 0.704 $\pm$ 0.034 & \textbf{0.823 $\pm$ 0.032} & 0.0008* \\
& \mbox{Precision (S)}  & 0.585 $\pm$ 0.023 & \textbf{0.708 $\pm$ 0.044} & 0.0021* \\
& \mbox{Recall (S)}     & 0.954 $\pm$ 0.042 & 0.846 $\pm$ 0.094          & -- \\
& \mbox{F1-Score} (S)   & 0.725 $\pm$ 0.024 & \textbf{0.768 $\pm$ 0.045} & 0.0655 \\
& Accuracy       & 0.715 $\pm$ 0.027 & \textbf{0.800 $\pm$ 0.035} & 0.0043* \\
& AUC            & 0.908 $\pm$ 0.006 & \textbf{0.914 $\pm$ 0.036} & 0.3872 \\
\midrule

\multirow{8}{*}{CIP}
& \mbox{Precision (R)}  & 0.915 $\pm$ 0.040 & 0.914 $\pm$ 0.003          & -- \\
& \mbox{Recall (R)}     & 0.873 $\pm$ 0.093 & \textbf{0.973 $\pm$ 0.041} & 0.0314* \\
& \mbox{F1-Score} (R)   & 0.891 $\pm$ 0.058 & \textbf{0.942 $\pm$ 0.021} & 0.0591 \\
& \mbox{Precision (S)}  & 0.867 $\pm$ 0.088 & \textbf{0.968 $\pm$ 0.047} & 0.0313* \\
& \mbox{Recall (S)}     & 0.905 $\pm$ 0.044 & 0.895 $\pm$ 0.000          & -- \\
& \mbox{F1-Score} (S)   & 0.884 $\pm$ 0.054 & \textbf{0.929 $\pm$ 0.022} & 0.0729 \\
& Accuracy       & 0.888 $\pm$ 0.056 & \textbf{0.937 $\pm$ 0.022} & 0.0646 \\
& AUC            & 0.949 $\pm$ 0.022 & \textbf{0.951 $\pm$ 0.012} & 0.4555 \\
\bottomrule
\end{tabular}
\end{table*}

\begin{figure}[htbp]
    \centering
    \begin{subfigure}[b]{0.32\textwidth}
        \centering
        \includegraphics[width=\linewidth, trim=1.5cm 2.5cm 2cm 1.5cm, clip]{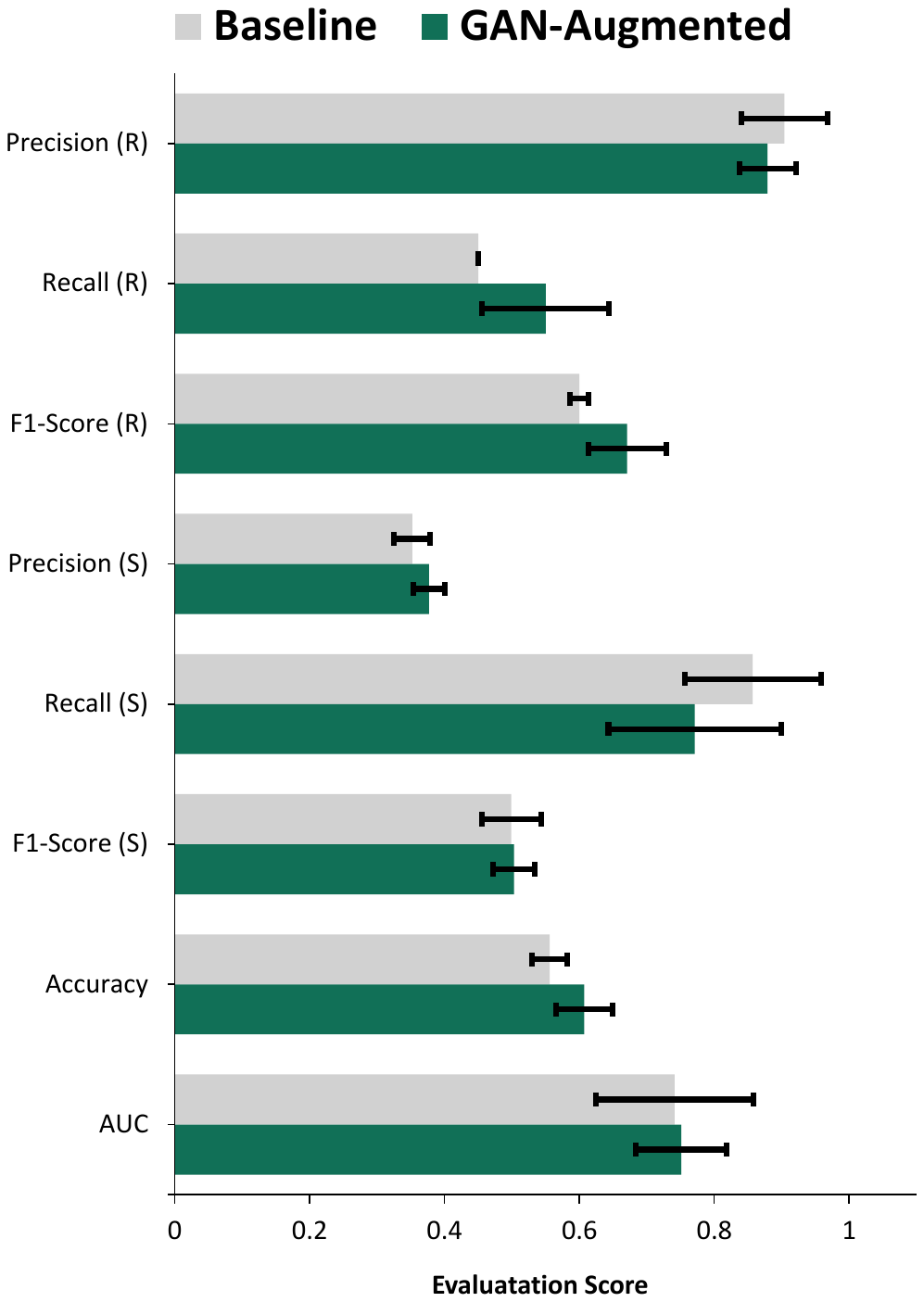}
        \caption{}
        \label{fig:chart_azi_aware}
    \end{subfigure}
    \begin{subfigure}[b]{0.32\textwidth}
        \centering
        \includegraphics[width=\linewidth, trim=1.5cm 2.5cm 2cm 1.5cm, clip]{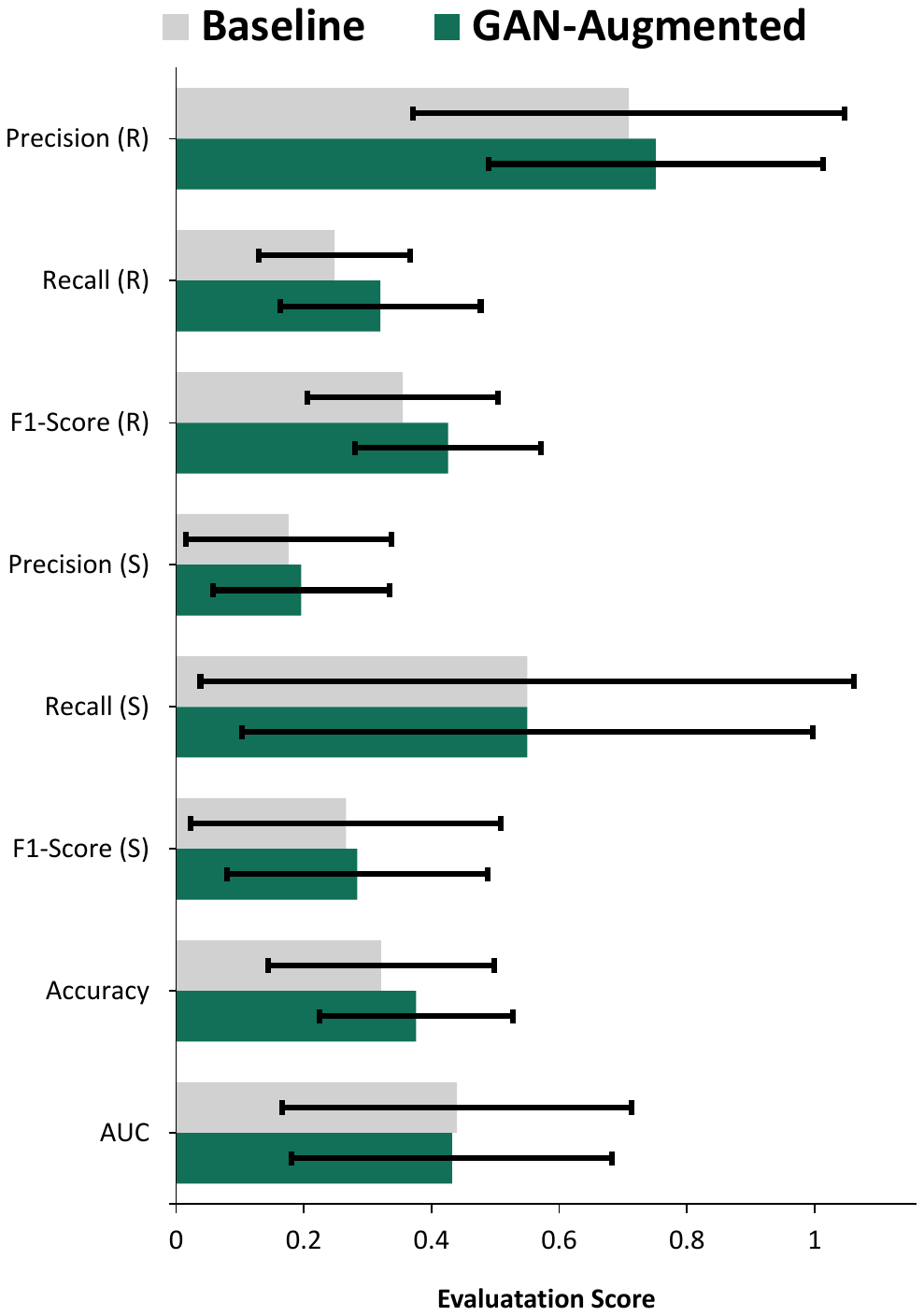}
        \caption{}
        \label{fig:chart_TE_aware}
    \end{subfigure}
    \begin{subfigure}[b]{0.32\textwidth}
        \centering
        \includegraphics[width=\linewidth, trim=1.5cm 2.5cm 2cm 1.5cm, clip]{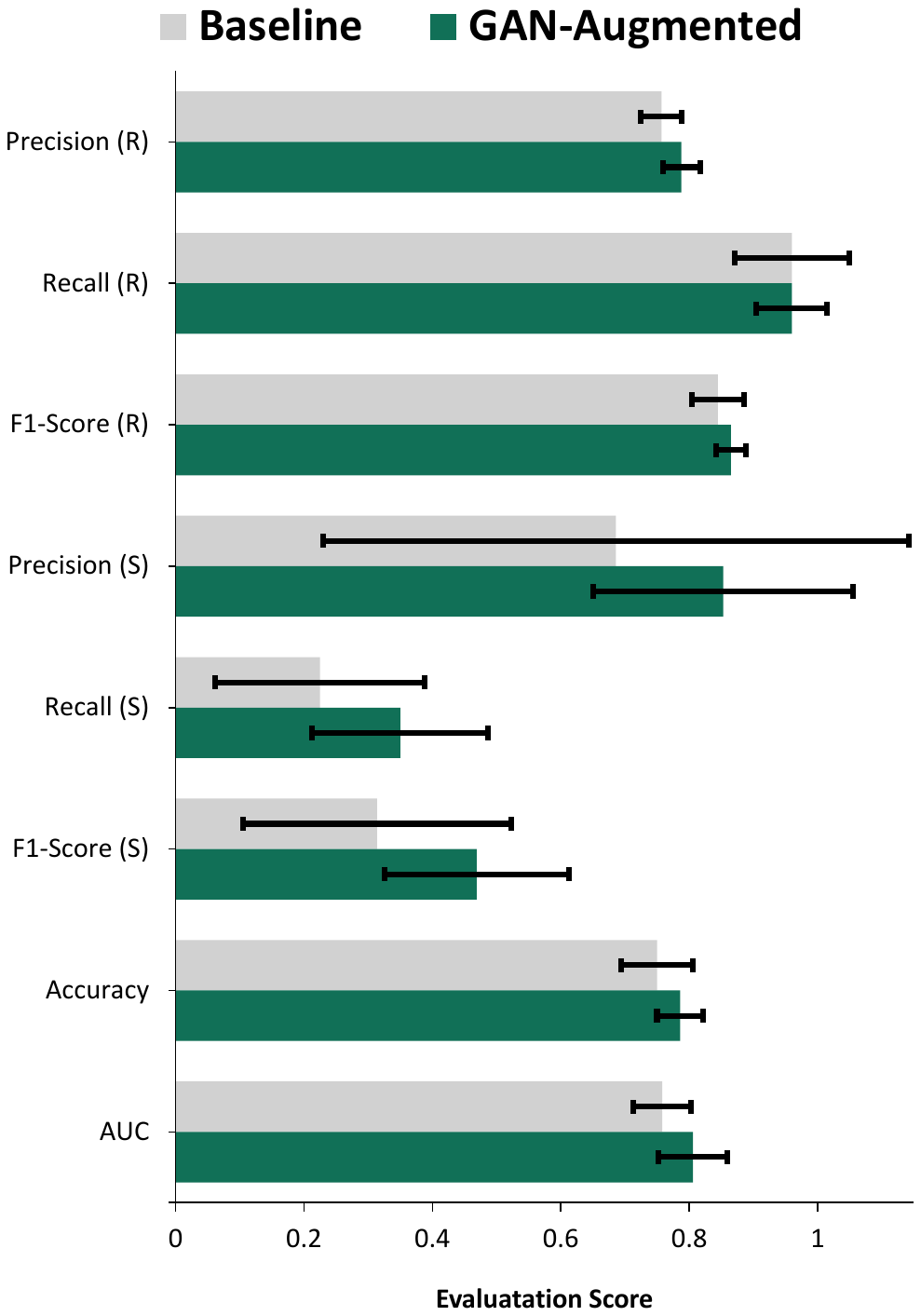}
        \caption{}
        \label{fig:chart_cip_aware}
    \end{subfigure}
    \caption{(Isolate-Aware Splitting) Performance comparison between Baseline and GAN-Augmented training for (a) AZI, (b) TE, and (c) CIP.}
    \label{fig:chart_aware}
\end{figure}

% \subsubsection{Isolate-Aware Splitting}
Table~\ref{tab:metrics_isolate} and Figure~\ref{fig:chart_aware} report the results when test isolates are never seen during training.
This is the more demanding setting: absolute performance is markedly lower and more variable across seeds than under isolate-agnostic splitting.
Baseline accuracy is only $0.556$ (AZI), $0.321$ (TE) and $0.750$ (CIP).
The large standard deviations (e.g., \mbox{Recall (S)} of $0.550 \pm 0.512$ for TE) are consistent with the small number of test isolates, where a change in the prediction for a single image shifts the metrics substantially.

Despite this difficulty, GAN augmentation yielded higher mean Accuracy for all three antibiotics ($0.051$ for AZI, $0.055$ for TE, and $0.036$ for CIP), although none of these differences reached significance. 
Statistically significant improvements ($p<0.05$) were found in two metrics for AZI (\mbox{Recall (R)}, \mbox{F1-Score} (R)), one for TE (\mbox{Recall (R)}), and four for CIP (\mbox{Precision (R)}, \mbox{Recall (S)}, \mbox{F1-Score} (S), AUC).

For AZI and TE, the baseline detects few resistant isolates (\mbox{Recall (R)} of $0.450$ and $0.248$), and augmentation significantly raises this recall (to 0.550, $p=0.038$ for AZI, and to $0.320$, $p=0.044$ for TE). 
This improvement is potentially clinically relevant because a resistant isolate misclassified as susceptible could lead to inappropriate antibiotic treatment. 
For CIP, the baseline almost always predicts resistant (\mbox{Recall (R)} $=0.960$) but rarely identifies susceptible isolates (\mbox{Recall (S)} $=0.225$), and augmentation mainly improves the susceptible class (\mbox{Recall (S)}: from $0.225$ to $0.350$; $p=0.045$, \mbox{F1-Score} (S): from $0.314$ to $0.469$, $p=0.041$), while keeping \mbox{Recall (R)} unchanged at $0.960$.  CIP is the only antibiotic for which the threshold-independent AUC improved significantly (from $0.758$ to $0.806$, $p=0.023$).

In other words, augmentation tends to improve the class the baseline handles worst. The one visible cost is for AZI, where \mbox{Precision (R)} and \mbox{Recall (S)} decrease slightly ($0.025$ and $0.086$, not significant).

\begin{table*}[htbp]
\centering
\caption{(Isolate-Aware Splitting) Per-antibiotic comparison between Baseline and GAN-Augmented training. Image-level hard voting, mean $\pm$ std over 5 seeds. \textbf{Bold}: numerically higher score within each metric for the same antibiotic. $* = p < 0.05$ (paired one-sided $t$-test, $H_1$: Baseline $<$ GAN-Augmented). }
\label{tab:metrics_isolate}

\renewcommand{\arraystretch}{1.15}
\begin{tabular}{c l c c c}
\toprule
\textbf{Antibiotic} & \textbf{Metric} & \textbf{Baseline} & \textbf{GAN-Augmented} & \textbf{$p$-value} \\
\midrule

\multirow{8}{*}{AZI} 
& \mbox{Precision (R)}    & 0.904 $\pm$ 0.064 & 0.879 $\pm$ 0.042          & -- \\
& \mbox{Recall (R)}       & 0.450 $\pm$ 0.000 & \textbf{0.550 $\pm$ 0.094} & 0.038* \\
& \mbox{F1-Score} (R)     & 0.600 $\pm$ 0.014 & \textbf{0.671 $\pm$ 0.058} & 0.045* \\
& \mbox{Precision (S)}    & 0.352 $\pm$ 0.027 & \textbf{0.377 $\pm$ 0.023} & 0.147 \\
& \mbox{Recall (S)}       & 0.857 $\pm$ 0.101 & 0.771 $\pm$ 0.128          & -- \\
& \mbox{F1-Score} (S)     & 0.499 $\pm$ 0.044 & \textbf{0.503 $\pm$ 0.031} & 0.428 \\
& Accuracy         & 0.556 $\pm$ 0.026 & \textbf{0.607 $\pm$ 0.042} & 0.080 \\
& AUC              & 0.741 $\pm$ 0.117 & \textbf{0.751 $\pm$ 0.067} & 0.445 \\
\midrule
 
\multirow{8}{*}{TE}
& \mbox{Precision (R)}    & 0.709 $\pm$ 0.338 & \textbf{0.751 $\pm$ 0.262} & 0.441 \\
& \mbox{Recall (R)}       & 0.248 $\pm$ 0.118 & \textbf{0.320 $\pm$ 0.157} & 0.044* \\
& \mbox{F1-Score} (R)     & 0.355 $\pm$ 0.149 & \textbf{0.426 $\pm$ 0.146} & 0.107 \\
& \mbox{Precision (S)}    & 0.176 $\pm$ 0.161 & \textbf{0.196 $\pm$ 0.138} & 0.441 \\
& \mbox{Recall (S)}       & 0.550 $\pm$ 0.512 & 0.550 $\pm$ 0.447          & -- \\
& \mbox{F1-Score} (S)     & 0.266 $\pm$ 0.243 & \textbf{0.284 $\pm$ 0.204} & 0.464 \\
& Accuracy         & 0.321 $\pm$ 0.177 & \textbf{0.376 $\pm$ 0.152} & 0.320 \\
& AUC              & 0.440 $\pm$ 0.274 & 0.432 $\pm$ 0.251          & -- \\
\midrule

\multirow{8}{*}{CIP}
& \mbox{Precision (R)}    & 0.757 $\pm$ 0.032 & \textbf{0.788 $\pm$ 0.029} & 0.040* \\
& \mbox{Recall (R)}       & 0.960 $\pm$ 0.089 & 0.960 $\pm$ 0.055          & -- \\
& \mbox{F1-Score} (R)     & 0.845 $\pm$ 0.041 & \textbf{0.865 $\pm$ 0.023} & 0.249 \\
& \mbox{Precision (S)}    & 0.686 $\pm$ 0.456 & \textbf{0.853 $\pm$ 0.202} & 0.284 \\
& \mbox{Recall (S)}       & 0.225 $\pm$ 0.163 & \textbf{0.350 $\pm$ 0.137} & 0.045* \\
& \mbox{F1-Score} (S)     & 0.314 $\pm$ 0.209 & \textbf{0.469 $\pm$ 0.144} & 0.041* \\
& Accuracy         & 0.750 $\pm$ 0.056 & \textbf{0.786 $\pm$ 0.036} & 0.187 \\
& AUC              & 0.758 $\pm$ 0.045 & \textbf{0.806 $\pm$ 0.054} & 0.023* \\
\bottomrule
\end{tabular}
\end{table*}

\subsection{Discussion}
\label{sec:discussion}
Two findings are consistent across both strategies.
First, GAN augmentation significantly improved \mbox{Recall (R)} for AZI and TE under both strategies, and for CIP under the isolate-agnostic strategy, where the baseline had more room to improve (from $0.873$ to $0.973$, $p=0.031$).
Under isolate-aware splitting, CIP \mbox{Recall (R)} was already high at baseline ($0.960$) and remained unchanged.
Second, the gains are concentrated in the class that the baseline predicts least reliably.

%The strategies differ in magnitude and reliability.
Under isolate-agnostic splitting, improvements are larger and more uniform, but test isolates also appear in training, so the classifier may exploit isolate-specific appearance rather than resistance-related morphology.
Under isolate-aware splitting, performance drops sharply and improvements are smaller and drug-dependent.

Overall, GAN-based augmentation consistently improves the detection of resistant isolates, which is the clinically most important outcome, at the cost of a modest reduction in resistant-class precision and susceptible-class recall for some antibiotics.

There are still several limitations. First, the augmentation ratio is fixed at 50\% of the baseline training set; whether smaller (e.g., 25\%) or larger (e.g., 100\%) ratios would yield better or worse results remains untested. Second, the GAN was trained on the same dataset as the classifiers, sharing the same image-level train/validation/test splits. While cell-level separation ensures no direct leakage of test samples into GAN training, the generative model has implicitly learned the statistical distribution of the full training pool, which may inflate the apparent benefit of augmentation. Future work should validate this augmentation strategy on a held-out external dataset to provide a more unbiased estimate of its practical utility.

\section{Conclusion and Future Works}
\label{sec:conclusion}

We studied whether the morphology of an \textit{E.~coli} isolate under antibiotic
challenge can be synthesized from its genome, and formulated the task as
conditional image generation driven by a genomic foundation-model embedding. The
framework combines a genomic encoding pipeline, which reduces the raw paired-end
reads of an isolate to one 768-dimensional vector through a
segment--contig--isolate hierarchy with length-aware aggregation, and a
conditional Wasserstein GAN with gradient penalty that generates colony images
from that embedding together with the applied antibiotic and the resistance
status. Reaching a usable model required more than the adversarial objective
alone: a pixel-level reconstruction term was needed to escape the degenerate
all-black solution, and conditioning the critic on the full biological context
was needed to escape the subsequent collapse onto a single average colony.

Among seven ImageNet-pretrained backbones evaluated inside the generator under
identical conditions, ConvNeXt-Tiny gave the best distributional agreement with
real images (FID 80.88, KID 0.026), while SSIM stayed within a narrow band
(0.782--0.809) across all backbones and ranked EfficientNet-B0 and
InceptionNet-V1 highly despite their much worse FID and KID. We therefore treat
SSIM as a complementary reference rather than a selection criterion in this
setting.

The extrinsic evaluation is the more informative one. Adding
resistance-conditioned synthetic cells to a training set reduced to 25\% of its
real cells improved resistant-class recall significantly for all three
antibiotics when isolates were shared across splits (by 0.083 for AZI, 0.210 for
TE, and 0.100 for CIP), and accuracy rose for every antibiotic under both
splitting protocols. Under the stricter protocol, in which no test isolate is
seen during training, absolute performance fell sharply and the gains became
smaller and drug-dependent, with significant improvements in resistant-class
recall for AZI and TE and in the susceptible-class metrics and AUC for CIP.
Across both protocols the improvement concentrated in whichever class the
baseline handled worst, and it came with a modest loss of resistant-class
precision and susceptible-class recall for some drugs. For antimicrobial
susceptibility testing that trade-off is usually the preferable direction, since
a resistant isolate reported as susceptible carries a heavier clinical cost than
the converse, but the drop in the isolate-aware setting shows that
generalization to unseen isolates remains the binding constraint rather than
image fidelity.

Several limitations bound these conclusions and set the agenda for further work.
The dataset is a single species represented by 22 isolates from one specimen
type, which is the main reason the isolate-aware results are both weak and
high-variance; the large standard deviations there follow directly from the small
number of test isolates. The generator and the classifiers were trained on the
same dataset and the same image-level splits, so although cell-level separation
prevents direct leakage of test samples into generator training, the generative
model has seen the statistical distribution of the training pool and the apparent
benefit of augmentation may be optimistic. Validation on an external, independently
collected dataset is the single most important next step. The augmentation ratio
was fixed at 50\% of the retained real cells and was not swept, so the
relationship between the amount of synthetic data and downstream performance is
unknown. Images were generated at $64{\times}64$, which is sufficient for
single-cell appearance but discards detail that higher-resolution synthesis might
recover. Finally, the model is evaluated by what its samples do for a classifier
and not by whether it has captured any specific resistance mechanism; relating the
genomic embedding, or the generated morphology, to known resistance determinants
would turn the framework from a data-augmentation tool into an interpretable one.
Extending the panel beyond three antibiotics, enlarging the isolate cohort, and
testing whether a single multi-drug conditional model outperforms the
per-antibiotic classifiers used here are the natural continuations.

\section*{Acknowledgments}
This research is funded by Vietnam National University Ho Chi Minh City (VNU-HCM) under grant number A2025-18-02.

\bibliography{sn-bibliography}%

\end{document}